\documentclass[11pt]{article}
\usepackage[T1]{fontenc}
\usepackage{lmodern,microtype}
\usepackage[letterpaper,margin=1in]{geometry}
\usepackage{amsmath,amssymb,amsthm,mathtools,enumitem,booktabs}
\usepackage{algorithm}
\usepackage{algpseudocode}
\usepackage[numbers,sort&compress]{natbib}
\usepackage{xcolor}
\usepackage[colorlinks=true,linkcolor=black,citecolor=black,urlcolor=blue!55!black]{hyperref}
\newtheorem{theorem}{Theorem}[section]
\newtheorem{lemma}[theorem]{Lemma}
\newtheorem{proposition}[theorem]{Proposition}
\newtheorem{corollary}[theorem]{Corollary}
\theoremstyle{definition}
\newtheorem{definition}[theorem]{Definition}
\theoremstyle{remark}
\newtheorem{remark}[theorem]{Remark}
\setlist{topsep=4pt,itemsep=3pt,parsep=0pt}
\hypersetup{pdftitle={Breaking the Joint Action Space Dependence in Distributionally Robust Multi-Agent Reinforcement Learning},pdfauthor={Yue Wang}}
\algrenewcommand\algorithmicrequire{\textbf{Input:}}
\algrenewcommand\algorithmicensure{\textbf{Output:}}
\algrenewcommand\algorithmiccomment[1]{\hfill{\footnotesize\textit{$\triangleright$ #1}}}
\newcommand{\CEgap}{\operatorname{Gap}_{\mathrm{CCE}}}
\newcommand{\rob}{\mathcal{T}}                 
\newcommand{\spn}{\operatorname{sp}}           
\newcommand{\TV}{\operatorname{TV}}            
\newcommand{\E}{\mathbb{E}}
\newcommand{\Prob}{\mathbb{P}}
\newcommand{\cS}{\mathcal{S}}
\newcommand{\cA}{\mathcal{A}}
\newcommand{\cP}{\mathcal{P}}
\newcommand{\cF}{\mathcal{F}}

\newcommand{\one}{\mathbf{1}}                  
\newcommand{\ind}[1]{\mathbf{1}\{#1\}}         
\newcommand{\epsz}{\varepsilon_0}              

\title{Breaking the Joint Action Space Dependence in Distributionally Robust Multi-Agent Reinforcement Learning}
\author{Yue Wang \\ University of Central Florida}
\date{}
\begin{document}
\maketitle
\begin{abstract}
We study the sample complexity of learning robust coarse correlated equilibria (CCE) in finite-horizon, general-sum, distributionally robust Markov games with known rewards and a generative model of the nominal transitions. Each player evaluates a policy by its worst-case return over total-variation (TV) balls centered at the nominal transition distribution of every stage, state, and joint action; this is the standard $(s,a)$-rectangular uncertainty set. Existing generative-model guarantees for this model scale with the number $\prod_iA_i$ of joint actions, where $A_i$ is the number of actions of player~$i$. Since this number grows exponentially with the number of players, these guarantees suffer from the curse of multiagency. We design an algorithm whose sample complexity depends on the action sets only through the largest individual action-set size $A_*\coloneqq\max_iA_i$. The algorithm uses $\widetilde O(SA_*H^5D^4/\varepsilon^4)$ transition samples and, with probability at least $1-\delta$, returns a Markov policy that is an $\varepsilon$-robust CCE from every initial state. Here $S$ is the number of states, $H$ is the horizon, and $D\coloneqq\max_i\min\{H,1/\rho_i\}$, where $\rho_i$ is the TV radius of player~$i$. Two ingredients make this possible: a plug-in estimate of the robust backup whose mean error does not depend on the number of states, and a single batch of samples per round that serves as bandit feedback for all players at once. The result shows that the dependence on $\prod_iA_i$ can be avoided without modifying the $(s,a)$-rectangular uncertainty set.
\end{abstract}

\section{Introduction}\label{sec:intro}
Reinforcement learning (RL) policies are often trained in a simulator, or on data collected from a nominal environment, and then deployed in an environment whose dynamics differ from the nominal ones. Even small differences can severely degrade a policy that is optimal for the nominal model, a phenomenon known as the sim-to-real gap~\cite{zhao2020simtoreal}. One practical remedy is to train against adversarial disturbances~\cite{pinto2017robust}. A more principled alternative is the robust Markov decision process (MDP)~\cite{iyengar2005robust,nilim2005robust,wiesemann2013robust}, which evaluates a policy by its worst-case return over an uncertainty set of transition kernels around the nominal one. When this set is rectangular, that is, a product of sets indexed by state--action pairs, the robust values satisfy a dynamic program in which the worst case is taken row by row~\cite{iyengar2005robust,nilim2005robust}. Distributionally robust RL learns such policies from samples of the nominal environment. For a single agent with a generative model, its sample complexity has been characterized for several uncertainty sets~\cite{panaganti2022sample,yang2022toward}, and minimax-optimal rates are known for total-variation (TV) balls~\cite{shi2023curious}.

Many applications involve several decision makers whose actions jointly drive a shared environment, from multi-player games~\cite{vinyals2019grandmaster} to robotics and autonomous driving~\cite{zhang2021multi}. Markov games~\cite{shapley1953stochastic,littman1994markov} are the standard model for such problems, and their natural solution concepts are equilibria. Computing a Nash equilibrium (NE) is PPAD-hard already in two-player normal-form games~\cite{daskalakis2009complexity,chen2009settling}, and the number of samples needed to learn an NE of a general-sum Markov game can grow exponentially with the number of players~\cite{song2021can}. The correlated equilibrium (CE)~\cite{aumann1974subjectivity} and the coarse correlated equilibrium (CCE)~\cite{moulin1978strategically} relax the NE by letting a mediator correlate the players' actions. Both are defined by linear constraints, and both arise from simple learning dynamics. When every player has vanishing external regret, the empirical distribution of play approaches the set of CCE; vanishing internal regret yields CE~\cite{hart2000simple,cesa2006prediction}. For this reason, the CCE has become a standard target for learning in multi-player general-sum Markov games~\cite{jin2021v,song2021can,daskalakis2023complexity}.

Model mismatch is especially delicate when several players interact, because the equilibria of two nearby games can differ substantially; \citet{shi2024sample} give an example. Robust Markov games~\cite{kardes2011discounted,zhang2020robust} address this issue by letting every player evaluate a joint policy through its worst-case return over its own set of transition kernels (see~\cite{aghassi2006robust} for one-shot games with payoff uncertainty). Robust equilibria exist under standard conditions~\cite{kardes2011discounted,zhang2020robust}, but computing a robust NE is PPAD-hard even in two-player zero-sum games with reward uncertainty~\cite{mcmahan2024roping}. As in the nominal case, we therefore focus on the robust CCE, a joint recommendation policy under which no player can increase its worst-case return by unilaterally ignoring its recommendations~\cite{shi2024sample}. This paper asks how many samples are needed to learn a robust CCE and, in particular, whether this number must grow with the number $\prod_iA_i$ of joint actions, where $A_i$ is the number of actions of player~$i$. Because $\prod_iA_i$ is exponential in the number of players, such growth is known as the curse of multiagency~\cite{jin2021v,song2021can,shi2024breaking}.

For nominal Markov games, this question has largely been answered. V-learning~\cite{jin2021v,song2021can,mao2023provably} and later algorithms~\cite{daskalakis2023complexity,wang2023breaking,cui2023breaking,li2022minimax} learn CCE with sample complexities that grow polynomially with $\max_iA_i$ instead of $\prod_iA_i$ (Section~\ref{sec:related}). For robust Markov games, the picture is much less complete. In the standard $(s,a)$-rectangular model with TV balls, both the generative-model guarantee of \citet{shi2024sample} and the online regret bounds of \citet{farhat2026online} scale with $\prod_iA_i$. The results that avoid this factor either change the uncertainty model~\cite{shi2024breaking,jiao2024minimax} or rely on additional conditions. \citet{ma2023decentralized} bound the diameter of the uncertainty sets and assume an accurate estimator of the robust backup, and the offline bound of \citet{li2025sample} scales with a concentrability coefficient of the data that is defined over joint actions. We are not aware of a result that learns a robust CCE of the standard model with a sample complexity free of $\prod_iA_i$, for all radii and without coverage assumptions.

The difficulty comes from how the uncertainty sets are indexed. In the standard $(s,a)$-rectangular model of \citet{shi2024sample}, player~$i$ has a TV ball of radius $\rho_i$ around the nominal transition distribution $P_h(\cdot\mid s,a)$ of every stage $h$, state $s$, and \emph{joint} action $a$; we call these distributions the \emph{nominal rows}. The balls are taken in the full probability simplex, so an adversary may move mass to states that have zero nominal probability. For a row $p$, a radius $\rho$, and a vector $u$ of continuation values, the worst-case continuation value is the \emph{robust backup}
\[
 \rob_\rho(p,u)\coloneqq\min_{q:\,\TV(q,p)\le\rho}qu .
\]
The robust backup is convex in $p$ but in general not affine, and this nonlinearity is the source of the difficulty.

To see why, suppose that the other players randomize. The robust payoff of an action is then the \emph{average of the robust backups} over the realized joint actions, and by convexity this average can be strictly larger than the robust backup of the averaged row. For example, let $u$ take the values $0$ and $1$ on two states, let $\rho=1/2$, and let two equally likely joint actions lead deterministically to the first and to the second state, respectively. The two robust backups are $0$ and $1/2$, so their average is $1/4$, whereas the robust backup of the averaged row $(1/2,1/2)$ is $0$. Averaging the nominal rows before taking the robust backup would therefore change the payoff. The same example shows why a single successor state is not enough. The expectation of any statistic of one successor is affine in the row from which the successor is drawn, while the three rows above satisfy a strict convexity inequality for $p\mapsto\rob_\rho(p,u)$. Hence no statistic of a single successor is an unbiased estimate of the robust backup at every nominal row, yet the nominal algorithms rely on exactly this unbiasedness. The algorithm of \citet{shi2024sample} sidesteps the issue by estimating the row of each of the $\prod_iA_i$ joint actions at every stage and state, and its sample complexity scales accordingly.

\textbf{Main result.}
We show that the curse of multiagency can be avoided in this model. Consider an $m$-player general-sum Markov game with $S$ states, horizon $H\ge2$, known rewards in $[0,1]$, and $A_i$ actions for player~$i$, and suppose that the learner has access to a generative model of the nominal kernel. Let $A_*\coloneqq\max_iA_i$ and $D\coloneqq\max_i\min\{H,1/\rho_i\}$, with the convention $1/0=+\infty$. Our algorithm uses
\[
 \widetilde O\!\left(\frac{SA_*H^5D^4}{\varepsilon^4}\right)
\]
transition samples and, with probability at least $1-\delta$, returns a Markov policy that is an $\varepsilon$-robust CCE simultaneously from every initial state (Theorem~\ref{thm:simple-global}).

The bound depends on the action sets only through $A_*$ and on the number of players only logarithmically. It holds for arbitrary player-specific radii $\rho_i\in[0,1]$, general-sum rewards, and stage-dependent nominal kernels, and it leaves the $(s,a)$-rectangular uncertainty set unchanged. Table~\ref{tab:comparison} places the result among existing guarantees. Compared with the bound of \citet{shi2024sample} for the same model, ours replaces $\prod_iA_i$ by $A_*$ at the cost of larger powers of $H$, $D$, and $1/\varepsilon$. In the range where both theorems apply, our bound is the smaller one when $\prod_iA_i/A_*>H^2D^3/\varepsilon^2$, up to logarithmic factors (Remark~\ref{rem:same-model}).

\begin{table}[t]
\centering
\caption{Upper bounds on the number of nominal transition samples needed to learn an $\varepsilon$-robust equilibrium with a generative model, up to logarithmic factors and within the parameter range of each theorem. Here $A_*=\max_iA_i$. In the TV rows, $D=\min\{H,1/\min_i\rho_i\}$, which equals $\max_i\min\{H,1/\rho_i\}$. In the contamination row, $D=\min\{H,1/\theta\}$ for the common contamination level $\theta\in[0,1)$. The convention is $1/0=+\infty$.}
\label{tab:comparison}
\smallskip
{\small\setlength{\tabcolsep}{4pt}
\begin{tabular}{@{}llll@{}}
\toprule
Reference & Uncertainty set & Equilibrium & Sample complexity\\
\midrule
\citet{shi2024sample} & $(s,a)$-rectangular, TV & NE, CE, CCE & $SH^3D\prod_iA_i/\varepsilon^2$\\[2pt]
\citet{jiao2024minimax} & $(s,a)$-rectangular, contamination & CCE & $SH^3D\sum_iA_i/\varepsilon^2$\\[2pt]
\citet{shi2024breaking} & policy-induced $(s,a_i)$-rectangular, TV & CCE & $SH^6D\sum_iA_i/\varepsilon^4$\\[2pt]
This paper & $(s,a)$-rectangular, TV & CCE & $SA_*H^5D^4/\varepsilon^4$\\
\bottomrule
\end{tabular}}
\end{table}

\textbf{Overview of the approach.}
The algorithm works backward in time. When it processes a stage $h$ and a state $s$, the continuation values of stage $h+1$ are already fixed, so the players face a one-shot game whose payoffs are the rewards plus the robust backups of these values. The players learn an approximate CCE of this local game by repeated play. In each round, they draw a joint action from the product of their current mixed strategies, and the learner queries a batch of $n$ successor states from the single row of this joint action. Every player evaluates the robust backup of the \emph{empirical} distribution of the batch with its own continuation values and its own radius, and uses the result as bandit feedback for an exponential-weights update with implicit exploration~\cite{neu2015}. The time-averaged strategies form the recommendation law at $(h,s)$, and the time-averaged payoffs become the continuation values passed to stage $h-1$.

Two facts make this construction work. The first is that the plug-in backup has a mean error of at most $\spn(u)/(2\sqrt n)$ at every row, where $\spn(u)$ is the span of the continuation vector $u$ (Proposition~\ref{est:bias}). This bound does not involve the number of states, and because it holds row by row, the algorithm needs neither an estimate of each row in TV distance nor a union bound over the joint actions. The second is that one batch can serve as bandit feedback for all players at once. Given the past, the action of each player is independent of everything that determines its counterfactual payoffs, so implicit exploration bounds the deviation gains of all players simultaneously (Section~\ref{sec:overview}). Backward induction then propagates the local errors to the first stage. Since the number of rounds and the batch size both scale with the inverse square of the local accuracy, the sample complexity carries the fourth power of $1/\varepsilon$ (Remark~\ref{rem:rate}).


\subsection{Related work}\label{sec:related}

\textbf{Robust MDPs and distributionally robust RL.}
\citet{iyengar2005robust} and \citet{nilim2005robust} showed that robust MDPs with rectangular uncertainty sets admit a robust dynamic program, and \citet{wiesemann2013robust} extended the analysis to $s$-rectangular sets. When the nominal model must be learned, model-based algorithms estimate every row of the nominal kernel and solve the resulting empirical robust MDP. \citet{zhou2021finite} analyzed this approach for offline tabular RL with Kullback--Leibler (KL) uncertainty sets, and \citet{panaganti2022sample} and \citet{yang2022toward} derived generative-model sample complexities for TV, $\chi^2$, and KL uncertainty sets. \citet{shi2023curious} established minimax-optimal rates for TV balls over the full range of radii; their rates show that robust MDPs with TV uncertainty are never harder to learn than standard MDPs, whereas $\chi^2$ uncertainty can make learning considerably harder. Model-free methods include robust Q-learning and robust temporal-difference learning for contamination sets~\cite{wang2021online}, distributionally robust Q-learning with multilevel Monte Carlo estimates~\cite{liu2022distributionally,wang2023finite}, and policy-gradient methods with global optimality guarantees~\cite{wang2022policy}. Robust RL from offline data was studied in~\cite{panaganti2022robust,shi2024distributionally,blanchet2023double,wang2024unified}. With interactive data collection, \citet{lu2024distributionally} showed that sample-efficient learning is impossible without further assumptions, and they identified a vanishing-minimal-value condition under which TV-robust MDPs become learnable. All these methods are designed for the state--action pairs of a single agent. Applied directly to a Markov game, whose rows are indexed by joint actions, they would treat each of the $\prod_iA_i$ joint actions separately at every stage and state.

\textbf{Learning equilibria in nominal Markov games.}
Markov games were introduced by \citet{shapley1953stochastic} and proposed as a framework for multi-agent RL by \citet{littman1994markov}; \citet{zhang2021multi} survey the theory. For two-player zero-sum games, \citet{zhang2020model} gave a near-optimal model-based algorithm with a generative model, and \citet{bai2020provable}, \citet{bai2020near}, and \citet{liu2021sharp} developed sample-efficient self-play algorithms with online exploration. For multi-player general-sum games, the model-based guarantees of \citet{liu2021sharp} scale with $\prod_iA_i$~\cite{song2021can}. V-learning~\cite{jin2021v,song2021can,mao2023provably} was the first approach to learn (non-Markov) CCE and CE from trajectories with sample complexities that scale polynomially with $\max_iA_i$ instead of $\prod_iA_i$~\cite{jin2021v,song2021can}. The same kind of dependence was later obtained for Markov CCE~\cite{daskalakis2023complexity,wang2023breaking}, also in large state spaces with independent linear function approximation~\cite{cui2023breaking}. With a generative model, the Q-FTRL algorithm of \citet{li2022minimax} learns a CCE with $\widetilde O(SH^4\sum_iA_i/\varepsilon^2)$ samples. Like these methods, our algorithm runs a separate no-regret learner for each player. It differs in the feedback, which must be a batch rather than a single transition because the robust backup is nonlinear in the row.

\textbf{Robust Markov games.}
\citet{kardes2011discounted} introduced discounted robust stochastic games, in which every player takes a robust-optimization approach to uncertain game data, and proved that equilibria exist. Their model complements the robust games of \citet{aghassi2006robust}, which treat one-shot games with payoff uncertainty. \citet{zhang2020robust} formulated robust multi-agent RL as a robust Markov game, defined a robust NE, and proposed a Q-learning algorithm with convergence guarantees under certain conditions, together with an actor-critic variant. \citet{li2019robust} considered robustness to changes in the other agents' policies, rather than in the transition kernel, through a minimax extension of multi-agent deep deterministic policy gradient. \citet{mcmahan2024roping} related robust Markov games with $s$-rectangular uncertainty to regularized Markov games and showed that computing a robust NE is PPAD-hard even for two-player zero-sum games with reward uncertainty. For cooperative agents, \citet{wang2022data} developed a decentralized robust multi-agent Q-learning algorithm with a sample complexity analysis, and \citet{roch2026distributionally} studied distributionally robust Markov games under the average-reward criterion.

\textbf{Sample complexity of general-sum robust Markov games.}
With a generative model, \citet[Theorem~4.1]{shi2024sample} estimate each nominal row from a fixed number of samples and run robust value iteration. For positive radii, their algorithm learns a robust NE, CE, or CCE with $\widetilde O(SH^3D\prod_iA_i/\varepsilon^2)$ samples. Our algorithm instead chooses the queried rows adaptively, from the players' current strategies; Remark~\ref{rem:same-model} compares the two upper bounds with the lower bound in their Theorem~4.2. For offline data, \citet{blanchet2023double} extended their framework for robust offline RL to robust Markov games. \citet{li2025sample} studied robust two-player zero-sum Markov games with TV uncertainty sets indexed by joint actions. They also extended their algorithm to robust NE of multi-player general-sum games, with a bound that scales with $\sum_iA_i$ times a robust unilateral concentrability coefficient of the dataset. This coefficient is defined over joint actions and measures how well the data cover unilateral deviations from the equilibrium.

For online interaction, \citet{ma2023decentralized} give a decentralized robust V-learning algorithm for uncertainty sets indexed by joint actions. Its episode complexity depends on $A_*$, and its output is a non-Markov robust correlated equilibrium. Their Theorems~12 and~13 require an accurate estimator of the robust backup and a bound on the diameter of the uncertainty sets. Theorem~12 restricts the diameter in terms of the target accuracy, and Theorem~13 bounds it by $p_{\min}/H$, where $p_{\min}$ is a lower bound on the state-visitation probabilities. \citet{farhat2026online} study robust Markov games with TV and KL uncertainty sets under online interaction, where the learner must also reach informative states along trajectories. Their regret bounds for NE, CE, and CCE scale with $\prod_iA_i$. The accompanying lower bound of order $\sqrt{K\prod_iA_i}$ over $K$ episodes concerns the regret with respect to robust NE, for which an exponential dependence on the number of players is expected already in nominal games~\cite{song2021can}. \citet{zheng2026distributionally} study online robust Markov games with linear function approximation. We work with a generative model instead, which removes the exploration problem and isolates the effect of joint-action uncertainty on the sample complexity.

\textbf{Robust formulations that avoid the curse of multiagency.}
Two formulations avoid the curse by changing the uncertainty model. \citet{shi2024breaking} center the TV ball of player~$i$ at the nominal kernel averaged over the other players' actions; for a correlated policy, the average is taken under the conditional distribution of these actions given player~$i$'s own action. The resulting uncertainty set, which they call fictitious, depends on the policy and is rectangular over $(s,a_i)$, and their CCE criterion averages robust values over a distribution of product policies. For positive radii, their sample complexity is $\widetilde O(SH^6D\sum_iA_i/\varepsilon^4)$, and \citet{gai2026} extend this formulation to linear function approximation. \citet{jiao2024minimax} keep the $(s,a)$-rectangular structure but replace the TV ball by a contamination set with a common contamination level $\theta$. The robust backup then equals $(1-\theta)\,pu+\theta\min_su(s)$, and their Theorem~1 gives $\widetilde O(SH^3\min\{H,1/\theta\}\sum_iA_i/\varepsilon^2)$ samples. In the first model, the robust backup is applied to the averaged row by definition; in the second, the backup is affine in $p$ and therefore commutes with averaging. The example above shows that neither simplification is available for TV balls around the individual rows.

\textbf{No-regret learning with bandit feedback.}
The local procedure of our algorithm is an instance of uncoupled no-regret dynamics, whose time-averaged play approaches the set of CCE~\cite{hart2000simple,cesa2006prediction}. Each player faces an adversarial bandit problem~\cite{auer2002nonstochastic}, and we use exponential weights with implicit exploration~\cite{kocak2014efficient,neu2015}, which admits high-probability regret bounds without explicit exploration. Our setting differs from the standard bandit setting in two respects. First, the feedback of all players is generated by one shared batch from the row of the realized joint action. Second, every loss is the plug-in robust backup of this batch, so the loss vector contains randomness that is generated in the current round. We handle the second point by applying the high-probability analysis of implicit exploration with respect to an enlarged filtration (Lemma~\ref{loc:ix}).

\textbf{Estimating nonlinear robust backups.}
Because the robust backup is a nonlinear functional of the transition row, plug-in estimates are biased. Model-based methods estimate every row and control the error of the resulting empirical robust Bellman operator~\cite{panaganti2022sample,yang2022toward,shi2023curious}. Multilevel Monte Carlo randomization~\cite{blanchet2015unbiased} produces unbiased estimators of nonlinear functions of expectations and has been used in distributionally robust Q-learning~\cite{liu2022distributionally,wang2023finite}. These estimators use a random number of samples. In~\cite{liu2022distributionally}, the expected number of samples is infinite, and \citet{wang2023finite} modified the estimator to make it finite. We take a different route. Every round uses a batch of fixed size, and the tail-integral form of the TV backup bounds its bias independently of the number of states (Proposition~\ref{est:bias}).

\section{Problem formulation}\label{sec:model}

\textbf{Notation.}
We write $[k]\coloneqq\{1,\dots,k\}$ and $x_+\coloneqq\max\{x,0\}$. For a finite set $\mathcal X$, $\Delta(\mathcal X)$ is the set of probability distributions on $\mathcal X$. For a vector $u\in\mathbb R^{\mathcal X}$ and a distribution $q\in\Delta(\mathcal X)$, we write $qu\coloneqq\sum_{x\in\mathcal X}q(x)u(x)$, $u_{\min}\coloneqq\min_xu(x)$, and $\spn(u)\coloneqq\max_xu(x)-u_{\min}$ for the span of $u$. The total-variation distance between $q,p\in\Delta(\mathcal X)$ is $\TV(q,p)\coloneqq\frac12\|q-p\|_1$. Inequalities between vectors hold componentwise, and $\one$ is the all-ones vector. The notation $\widetilde O(\cdot)$ hides absolute constants and factors that are logarithmic in $m$, $S$, $A_*$, $H$, and $1/\delta$.

\subsection{Robust Markov games with a generative model}

We consider a finite-horizon general-sum Markov game with $m\ge1$ players, a finite state space $\cS$ with $S\coloneqq|\cS|$ states, and $H$ stages. Player $i\in[m]$ has a finite action set $\cA_i$ with $A_i\coloneqq|\cA_i|$ actions. We write $\cA\coloneqq\prod_{i=1}^m\cA_i$ for the set of joint actions and $A_*\coloneqq\max_iA_i$. For a joint action $a\in\cA$, $a_i$ is the action of player~$i$, $a_{-i}$ collects the actions of the other players, and $(b,a_{-i})$ is the joint action obtained from $a$ by replacing $a_i$ with $b\in\cA_i$. At stage $h\in[H]$, in state $s$ and under the joint action $a$, player~$i$ receives a known deterministic reward $r_{i,h}(s,a)\in[0,1]$, and the next state is drawn from the nominal distribution $P_h(\cdot\mid s,a)\in\Delta(\cS)$. The nominal kernel $P\coloneqq\{P_h\}_{h\in[H]}$ may depend on the stage. We call a triple $(h,s,a)$, and by extension the distribution $P_h(\cdot\mid s,a)$, a \emph{nominal row}.

The learner does not know $P$, but it can sample from $P$ through a generative model. It may query any nominal row $(h,s,a)$, and given the transcript of earlier queries and the selected row, the simulator returns a successor state drawn from $P_h(\cdot\mid s,a)$ with fresh randomness. The rows may be chosen adaptively, that is, as functions of earlier responses and of the learner's internal randomness. The \emph{sample complexity} of an algorithm is its total number of queries. Since the rewards are known, computations that involve only the reward table cost no samples.

Each player guards against errors in the nominal kernel through its own uncertainty set. Player~$i$ has a known radius $\rho_i\in[0,1]$ and regards as possible every transition distribution within TV distance $\rho_i$ of the nominal row:
\[
 \cP_{i,h}(s,a)\coloneqq\bigl\{q\in\Delta(\cS):\TV\bigl(q,P_h(\cdot\mid s,a)\bigr)\le\rho_i\bigr\}.
\]
The uncertainty set of player~$i$ is the product $\cP_i\coloneqq\bigotimes_{(h,s,a)}\cP_{i,h}(s,a)$ over all nominal rows. It is rectangular over the triples $(h,s,a)$ with $a$ the \emph{joint} action, so the adversarial row may depend on the actions of all players; this is the agent-wise $(s,a)$-rectangularity of~\cite{shi2024sample}. Each player evaluates its own return, so the worst-case kernels, like the radii, may differ across players. We define
\[
 D_i\coloneqq\min\{H,1/\rho_i\},\qquad D\coloneqq\max_{i\in[m]}D_i ,
\]
with the convention $1/0=+\infty$, so that $D_i=H$ when $\rho_i=0$. Equivalently, $D=\min\{H,1/\min_i\rho_i\}$, and $1\le D_i\le D\le H$.

\subsection{Policies, robust values, and best responses}

An equilibrium compares the robust value of following a joint policy with the robust value of the best unilateral deviation from it. We define these objects in turn.

A \emph{Markov joint policy} $\pi=\{\pi_h(\cdot\mid s)\}_{h\in[H],s\in\cS}$ assigns a distribution $\pi_h(\cdot\mid s)\in\Delta(\cA)$ over joint actions to every stage and state. We interpret $\pi$ as a correlation device: at every visit to a pair $(h,s)$, the device draws $a\sim\pi_h(\cdot\mid s)$ with fresh randomness and privately recommends $a_i$ to player~$i$. The recommendations to different players may be correlated. We write $\pi_{-i,h}(\cdot\mid s)$ for the marginal distribution of $a_{-i}$ under $\pi_h(\cdot\mid s)$.

Suppose first that all players obey. For a kernel $P'\in\cP_i$, let $V^{\pi,P'}_{i,h}(s)$ be the expected total reward of player~$i$ from stage $h$ to stage $H$ when the game is in state $s$ at stage $h$, all players follow the recommendations of $\pi$, and the transitions follow $P'$. The robust value of player~$i$ is the worst case over its uncertainty set,
\begin{equation}\label{eq:robust-value}
 V^{\pi}_{i,h}(s)\coloneqq\min_{P'\in\cP_i}V^{\pi,P'}_{i,h}(s);
\end{equation}
Lemma~\ref{lem:bellman} below shows that the minimum is attained.

Player~$i$ may instead ignore its recommendations and follow a \emph{coarse deviation} $\sigma_i$. Such a policy selects the action $a_{i,k}$ at each stage $k$ as a possibly randomized function of the current state and of the states and joint actions observed before stage $k$; it does not see the recommendation issued at stage $k$. We write $\sigma_i\times\pi_{-i}$ for the behavior in which player~$i$ uses $\sigma_i$ and the other players obey $\pi$. Because the device draws each recommendation with fresh randomness, independent of the deviator's private randomization, the opponents' actions have law $\pi_{-i,k}(\cdot\mid s_k)$ given the history, the current state $s_k$, and the action chosen by $\sigma_i$. The best-response value of player~$i$ against $\pi_{-i}$ is
\begin{equation}\label{eq:br-value}
 V^{\dagger,\pi_{-i}}_{i,h}(s)
 \coloneqq\sup_{\sigma_i}\ \min_{P'\in\cP_i}V^{\sigma_i\times\pi_{-i},P'}_{i,h}(s),
\end{equation}
where the game starts in state $s$ at stage $h$ and the supremum is over all coarse deviations. (For each $\sigma_i$, the minimum is attained because $\cP_i$ is compact and the finite-horizon value is a polynomial in the entries of $P'$.)

Both values can be computed by backward induction, in which the worst case over kernels is taken row by row. The recursion uses the robust backup from Section~\ref{sec:intro}, written here for a continuation vector $u\in\mathbb R^{\cS}$, a row $p\in\Delta(\cS)$, and a radius $\rho\in[0,1]$:
\[
 \rob_\rho(p,u)\coloneqq\min_{q\in\Delta(\cS):\,\TV(q,p)\le\rho}qu .
\]

\begin{lemma}[Robust Bellman equations]\label{lem:bellman}
Let $\pi$ be a Markov joint policy and let $i\in[m]$. Set $V^{\pi}_{i,H+1}\coloneqq0$ and $V^{\dagger,\pi_{-i}}_{i,H+1}\coloneqq0$, and, for a given stage $h$ and state $s$, abbreviate $p_a\coloneqq P_h(\cdot\mid s,a)$. Then, for every $h\in[H]$ and $s\in\cS$,
\begin{align}
 V_{i,h}^{\pi}(s)
  &=\E_{a\sim\pi_h(\cdot\mid s)}
       \bigl[r_{i,h}(s,a)+\rob_{\rho_i}(p_a,V_{i,h+1}^{\pi})\bigr],\label{eq:obey}\\
 V_{i,h}^{\dagger,\pi_{-i}}(s)
  &=\max_{b\in\cA_i}\ \E_{a_{-i}\sim\pi_{-i,h}(\cdot\mid s)}
       \bigl[r_{i,h}(s,(b,a_{-i}))+\rob_{\rho_i}(p_{(b,a_{-i})},V_{i,h+1}^{\dagger,\pi_{-i}})\bigr].\label{eq:br}
\end{align}
Moreover, a single kernel attains the minimum in~\eqref{eq:robust-value} for all $(h,s)$, and a single deterministic Markov policy of player~$i$ attains the supremum in~\eqref{eq:br-value} for all $(h,s)$.
\end{lemma}

Lemma~\ref{lem:bellman} is an instance of robust dynamic programming for rectangular uncertainty sets~\cite{iyengar2005robust,nilim2005robust}, and \citet[equation~(13)]{shi2024sample} state the obedience equation~\eqref{eq:obey} in the present multi-agent form. They define the best-response value as a maximum over Markov deviations, whereas~\eqref{eq:br-value} allows history-dependent deviations against kernels fixed in advance. For this reason, Appendix~\ref{app:bellman} gives a short self-contained proof of both equations. The proof also shows that both values remain unchanged if nature may choose an admissible row after observing the history and the current joint action.

\subsection{Robust coarse correlated equilibrium and the learning goal}

\begin{definition}[Robust CCE]\label{def:cce}
The robust CCE gap of a Markov joint policy $\pi$ at an initial state $s$ is
\[
 \CEgap(\pi,s)\coloneqq\max_{i\in[m]}\bigl(V_{i,1}^{\dagger,\pi_{-i}}(s)-V_{i,1}^\pi(s)\bigr)_+ .
\]
For $\varepsilon\ge0$, the policy $\pi$ is an \emph{$\varepsilon$-robust CCE} if $\max_{s\in\cS}\CEgap(\pi,s)\le\varepsilon$.
\end{definition}

The positive part only makes the gap nonnegative. For a correlated policy, the difference inside the parentheses can be negative, because a coarse deviation cannot use the recommendation; a negative difference means that the player has no profitable deviation. Since the criterion takes the maximum over all initial states, an $\varepsilon$-robust CCE is approximately stable wherever the game starts. Finally, by Lemma~\ref{lem:bellman}, restricting the deviations to Markov policies does not change the best-response value, so our criterion coincides with the robust CCE of \citet{shi2024sample}, which also imposes the equilibrium inequality at every state.

Our goal is to learn such a policy from samples. Given an accuracy $\varepsilon\in(0,1]$ and a confidence level $\delta\in(0,1)$, the learner must output a Markov joint policy that is an $\varepsilon$-robust CCE with probability at least $1-\delta$. The probability is taken over the samples and the learner's internal randomness. We bound the number of queries that suffices for this purpose.

\section{Algorithm and main result}\label{sec:global}

The algorithm proceeds by backward induction over the stages. When it reaches stage $h$, it holds, for every player~$i$, a vector $U_{i,h+1}\in\mathbb R^{\cS}$ that estimates the robust values of this player at stage $h+1$ under the policy learned so far. At each state $s$, it then learns two objects: a recommendation law $\pi_h(\cdot\mid s)$ and one scalar $U_{i,h}(s)$ per player, which serves as a continuation value at stage $h-1$. Section~\ref{sec:alg-local} describes the procedure for a single stage and state, Section~\ref{sec:alg-backward} describes the backward recursion, and Section~\ref{sec:main-result} states the guarantee.

\subsection{The local procedure}\label{sec:alg-local}

Fix a stage $h$ and a state $s$, and regard the continuation vectors $v_i\coloneqq U_{i,h+1}$ as given. Abbreviate $r_i(a)\coloneqq r_{i,h}(s,a)$ and $p_a\coloneqq P_h(\cdot\mid s,a)$. The players then face a one-shot game, which we call the \emph{local game}, with payoffs
\begin{equation}\label{alg:local-game}
 g_i(a)\coloneqq r_i(a)+\rob_{\rho_i}(p_a,v_i),\qquad a\in\cA,\ i\in[m].
\end{equation}
The backward recursion needs an approximate CCE of the local game together with an estimate of every player's payoff under it. Algorithm~\ref{alg:local} learns both by repeated play. In each round $t$, every player~$i$ holds a mixed strategy $x_{i,t}$ over its own actions, and a joint action $a_t$ is drawn from the product distribution $x_t\coloneqq\bigotimes_ix_{i,t}$. The learner queries the row $p_{a_t}$ exactly $n$ times and forms the empirical distribution $\widehat p_t$ of the $n$ successors. Player~$i$ then computes the plug-in payoff $\widehat g_{i,t}\coloneqq r_i(a_t)+\rob_{\rho_i}(\widehat p_t,v_i)$, converts it into a loss in $[0,1]$, and updates the weight of the action it played by the exponential-weights rule with implicit exploration (EXP3-IX) of \citet{neu2015}.

\begin{algorithm}[!htb]
\caption{Shared-batch EXP3-IX for one local game}\label{alg:local}
\begin{algorithmic}[1]
\Require rewards $r_i:\cA\to[0,1]$, radii $\rho_i$, and continuation vectors $v_i\in\mathbb R^{\cS}$ for $i\in[m]$; query access to the rows $p_a$, $a\in\cA$; range $R$; number of rounds $T$; batch size $n$; parameter $\Lambda$
\State $c_i\gets\min_sv_i(s)$,\quad $\gamma_i\gets\sqrt{\Lambda/(A_iT)}$,\quad $w_{i,1}(b)\gets1$ \quad for all $i\in[m]$ and $b\in\cA_i$
\For{$t=1,\dots,T$}
  \State $x_{i,t}(b)\gets w_{i,t}(b)\big/\sum_{b'\in\cA_i}w_{i,t}(b')$ for all $i\in[m]$ and $b\in\cA_i$ \Comment{mixed strategies}
  \State draw $a_{i,t}\sim x_{i,t}$ independently for $i\in[m]$ and set $a_t\gets(a_{1,t},\dots,a_{m,t})$
  \State query the row $p_{a_t}$ $n$ times; $\widehat p_t\gets$ empirical distribution of the $n$ successors \Comment{shared batch}
  \For{each player $i\in[m]$}
    \State $\widehat g_{i,t}\gets r_i(a_t)+\rob_{\rho_i}(\widehat p_t,v_i)$ \Comment{observed payoff}
    \State $\ell_{i,t}\gets1-(\widehat g_{i,t}-c_i)/R$ \Comment{loss in $[0,1]$}
    \State $w_{i,t+1}(b)\gets w_{i,t}(b)\exp\Bigl\{-\gamma_i\,\dfrac{\ell_{i,t}\,\ind{a_{i,t}=b}}{x_{i,t}(b)+\gamma_i}\Bigr\}$ for all $b\in\cA_i$ \Comment{EXP3-IX update}
  \EndFor
\EndFor
\Ensure the joint distribution $\mu\coloneqq\frac1T\sum_{t=1}^Tx_{1,t}\otimes\cdots\otimes x_{m,t}$ and the estimates $U_i\coloneqq\frac1T\sum_{t=1}^T\widehat g_{i,t}$, $i\in[m]$
\end{algorithmic}
\end{algorithm}

We highlight three features of Algorithm~\ref{alg:local}.

\emph{One batch serves all players.} The batch in line~5 is drawn from a single row, and each player evaluates it with its own continuation vector $v_i$ and its own radius $\rho_i$. Each round therefore makes $n$ queries, regardless of the number of players.

\emph{Each player receives bandit feedback.} Player~$i$ observes only the plug-in payoff of the joint action that was played. The update in line~9 changes only the weight of the played action $a_{i,t}$, and its importance weight $1/(x_{i,t}(a_{i,t})+\gamma_i)$ involves only the player's own probability. The parameter $\gamma_i$ serves both as the learning rate and as the implicit-exploration parameter. The constants $c_i$ and $R$ only rescale the payoff to $[0,1]$: by Lemma~\ref{est:range}, $0\le r+\rob_{\rho_i}(p,v_i)-c_i\le1+(1-\rho_i)\spn(v_i)$ for every $r\in[0,1]$ and every distribution $p$, so $\ell_{i,t}\in[0,1]$ whenever $R\ge1+(1-\rho_i)\spn(v_i)$.

\emph{The output is a mixture of product distributions.} The law $\mu$ is the uniform mixture of the $T$ product distributions $x_t$. In general it is not a product distribution, and it differs from the product of the time-averaged marginals. To sample from $\mu$, one draws a round index $t$ uniformly from $[T]$, without revealing it to the players, and then draws each $a_i$ independently from $x_{i,t}$. Hence $\mu$ can be stored as the list of the vectors $x_{i,t}$, which takes $T\sum_iA_i$ numbers, and no table with $\prod_iA_i$ entries is ever formed.

\textbf{Computing the plug-in backup.}
Let $y_i\coloneqq v_i-c_i\one\ge0$, and let $X_{t,1},\dots,X_{t,n}$ be the successors observed in round $t$. Lemma~\ref{est:tail}, applied to the empirical distribution $\widehat p_t$, gives
\begin{equation}\label{alg:plugin}
 \rob_{\rho_i}(\widehat p_t,v_i)
 =c_i+\max_{0\le\alpha\le\spn(v_i)}
 \Bigl\{\frac1n\sum_{j=1}^n\min\{y_i(X_{t,j}),\alpha\}-\rho_i\alpha\Bigr\}.
\end{equation}
The objective is concave and piecewise linear in $\alpha$, its breakpoints are the observed values $y_i(X_{t,j})$, and it is nonincreasing for $\alpha\ge\max_jy_i(X_{t,j})$. Its maximum is therefore attained at one of the points $0,y_i(X_{t,1}),\dots,y_i(X_{t,n})$, and sorting the $n$ observed values computes the plug-in backup in $O(n\log n)$ time. The constant $c_i$ is the minimum of $v_i$ over the whole state space, including states that do not appear in the batch. This is how the estimate accounts for the full-simplex uncertainty set, in which the adversary may move mass to a state that the nominal row never visits.

\subsection{The backward recursion}\label{sec:alg-backward}

\textbf{Span bounds.}
The parameters of the local runs depend on the spans of the continuation vectors, so we need bounds on these spans that are available before any data are seen. For each player define
\begin{equation}\label{glob:range-recursion}
 d_{i,H+1}\coloneqq0,\qquad
 d_{i,h}\coloneqq1+(1-\rho_i)\,d_{i,h+1}
        =\sum_{k=0}^{H-h}(1-\rho_i)^k\quad(h\in[H]),
 \qquad d_h\coloneqq\max_{i\in[m]}d_{i,h},
\end{equation}
with the convention $0^0=1$. The recursion mirrors the algorithm: each stage adds a reward in $[0,1]$, and the robust backup shrinks the span of the continuation vector by the factor $1-\rho_i$ (Lemma~\ref{est:range}). Lemma~\ref{glob:range} shows that $\spn(U_{i,h})\le d_{i,h}$ on every run of the algorithm. The geometric sum in~\eqref{glob:range-recursion} has $H-h+1$ terms in $[0,1]$, the first of which equals one, and it is at most $1/\rho_i$ when $\rho_i>0$. Hence $1\le d_{i,h}\le\min\{H-h+1,1/\rho_i\}\le D_i$ for every $h\in[H]$.

\textbf{Parameters.}
Let $H\ge2$, $0<\varepsilon\le1$, and $0<\delta<1$. We set
\begin{equation}\label{glob:parameters}
\begin{gathered}
 \epsz\coloneqq\frac{\varepsilon}{2H},\qquad
 \Lambda\coloneqq\log\frac{8mA_*S(H-1)}{\delta},\\
 T_h\coloneqq\left\lceil\frac{256\,A_*d_h^2\Lambda}{\epsz^2}\right\rceil,\qquad
 n_h\coloneqq\left\lceil\frac{d_{h+1}^2}{\epsz^2}\right\rceil
 \qquad(h\in[H-1]).
\end{gathered}
\end{equation}
All local runs share the accuracy target $\epsz$, while their numbers of rounds $T_h$ and batch sizes $n_h$ grow with the span bounds $d_h$ and $d_{h+1}$. These bounds increase with the remaining horizon and saturate at $1/\min_i\rho_i$.

\begin{algorithm}[!htb]
\caption{Backward learning of a robust Markov CCE}\label{alg:backward}
\begin{algorithmic}[1]
\Require accuracy $\varepsilon$, confidence $\delta$, rewards $r_{i,h}$, radii $\rho_i$, query access to the nominal rows
\State set $U_{i,H+1}\gets0$ for all $i\in[m]$
\For{each state $s\in\cS$} \Comment{stage $H$: no queries}
  \State $\pi_H(\cdot\mid s)\gets$ an exact CCE of the one-shot game with payoffs $r_{i,H}(s,\cdot)$, $i\in[m]$
  \State $U_{i,H}(s)\gets\E_{a\sim\pi_H(\cdot\mid s)}r_{i,H}(s,a)$ for all $i\in[m]$
\EndFor
\If{$H=1$}
  \State \Return the policy $\pi=\{\pi_1(\cdot\mid s)\}_{s\in\cS}$
\EndIf
\State compute $d_h$ from~\eqref{glob:range-recursion} and $\epsz$, $\Lambda$, $T_h$, $n_h$ from~\eqref{glob:parameters}
\For{$h=H-1,\dots,1$}
  \For{each state $s\in\cS$}
    \State $(\mu,U_1,\dots,U_m)\gets$ output of Algorithm~\ref{alg:local} with rewards $r_{i,h}(s,\cdot)$, rows $P_h(\cdot\mid s,a)$,
    \Statex \hspace{\algorithmicindent}\hspace{\algorithmicindent}continuation vectors $v_i=U_{i,h+1}$, range $R=d_h$, $T=T_h$ rounds, batch size $n=n_h$, and $\Lambda$
    \State $\pi_h(\cdot\mid s)\gets\mu$ and $U_{i,h}(s)\gets U_i$ for all $i\in[m]$
  \EndFor
\EndFor
\Ensure the Markov joint policy $\pi=\{\pi_h(\cdot\mid s)\}_{h\in[H],s\in\cS}$
\end{algorithmic}
\end{algorithm}

\textbf{The recursion.}
Algorithm~\ref{alg:backward} states the complete procedure. At the last stage the continuation values are zero, so the local game at a state $s$ is the known reward game with payoffs $r_{i,H}(s,\cdot)$. The algorithm takes $\pi_H(\cdot\mid s)$ to be any exact CCE of this game, that is, any distribution $\nu\in\Delta(\cA)$ that satisfies the linear constraints
\begin{equation}\label{glob:terminal}
 \sum_{a\in\cA}\nu(a)\bigl[r_{i,H}(s,(b,a_{-i}))-r_{i,H}(s,a)\bigr]\le0
 \qquad\text{for all }i\in[m]\text{ and }b\in\cA_i.
\end{equation}
These constraints are feasible, because a Nash equilibrium of the reward game exists~\cite{nash_non-cooperative_1951} and its product distribution satisfies them. This step uses no query. At every earlier stage $h$, the vectors $U_{i,h+1}$ of all players are complete before any state of stage $h$ is processed, and Algorithm~\ref{alg:local} runs once per state, each time with fresh queries. At deployment, the device samples from the stored reward-game CCE at stage $H$ and from the stored mixture $\pi_h(\cdot\mid s)$ at each visit to $(h,s)$ with $h<H$, as described in Section~\ref{sec:alg-local}.

\subsection{Main result}\label{sec:main-result}

\begin{theorem}[Sample complexity of learning a robust Markov CCE]
\label{thm:simple-global}
Consider the robust Markov game of Section~\ref{sec:model} with arbitrary radii $\rho_i\in[0,1]$. Let $H\ge2$, $0<\varepsilon\le1$, and $0<\delta<1$, and run Algorithm~\ref{alg:backward} with the parameters in~\eqref{glob:parameters}.
\begin{enumerate}[label=\textup{(\roman*)}]
\item \textup{(Accuracy.)} With probability at least $1-\delta$, the output $\pi$ is an $\varepsilon$-robust CCE, that is,
\[
 \max_{s\in\cS}\CEgap(\pi,s)\le\varepsilon .
\]
\item \textup{(Sample complexity.)} The algorithm makes exactly
\begin{equation}\label{glob:exact-budget}
 N\coloneqq S\sum_{h=1}^{H-1}T_hn_h
\end{equation}
queries, and this deterministic number satisfies
\begin{equation}\label{glob:refined-budget}
 N\le C_0\,\frac{SA_*H^4\Lambda}{\varepsilon^4}
            \sum_{h=1}^{H-1}d_h^2d_{h+1}^2
 \le C_0\,\frac{SA_*H^5D^4}{\varepsilon^4}
       \log\frac{8mA_*SH}{\delta},
\end{equation}
where $C_0$ is an absolute constant; one may take $C_0=8224$.
\end{enumerate}
\end{theorem}

Appendix~\ref{app:global} proves Theorem~\ref{thm:simple-global}, and Section~\ref{sec:overview} outlines the argument. The count $N$ includes every sample used to evaluate a robust backup, and a single event covers all initial states. For $H=1$, Algorithm~\ref{alg:backward} makes no query and returns an exact robust CCE (Remark~\ref{rem:degenerate}). Three remarks explain the rate, compare it with the bound of \citet{shi2024sample}, and address computation.

\begin{remark}[Origin of the rate]\label{rem:rate}
A local run at stage $h$ uses $T_h=\widetilde O(A_*d_h^2/\epsz^2)$ rounds, the usual bandit rate for accuracy $\epsz$ and payoff range $d_h$. Each round uses a batch of $n_h=O(d_{h+1}^2/\epsz^2)$ samples, which brings the bias of the plug-in backup below $\epsz/2$. Their product contributes the factor $\epsz^{-4}=16H^4/\varepsilon^4$, and summing over the $S$ states and $H-1$ stages with $d_h\le D$ gives~\eqref{glob:refined-budget}. The batch is needed only because the robust backup is nonlinear in the row. If $\rho_i=0$ for every player, then $\rob_0(\widehat p,v_i)=\widehat pv_i$ is an unbiased estimate of $pv_i$, so the bias $\beta_i$ in Theorem~\ref{loc:certificate} vanishes even for a single sample. Theorem~\ref{loc:certificate} then gives both local bounds with error at most $\epsz$ for $n_h=1$ and $T_h$ as in~\eqref{glob:parameters}, and the proof of Theorem~\ref{thm:simple-global} applies without change. Since $d_h=H-h+1$ in this case, the transition count becomes
\[
 N=S\sum_{h=1}^{H-1}\left\lceil\frac{256A_*\Lambda(H-h+1)^2}{\epsz^2}\right\rceil
 =\widetilde O\!\left(\frac{SA_*H^5}{\varepsilon^2}\right),
\]
where we used $\sum_{j=2}^Hj^2\le H^3$ and $\epsz=\varepsilon/(2H)$. For positive radii, the batch multiplies the number of rounds by $n_h=O(H^2D^2/\varepsilon^2)$; this factor is the price that our analysis pays for the nonlinearity. A fourth power of $1/\varepsilon$ also appears in the guarantee of \citet{shi2024breaking} for policy-induced uncertainty sets, whose algorithm likewise draws a batch of samples in every iteration. Remark~\ref{est:dimension-free} shows that the bias of the plug-in estimator can indeed be of order $n^{-1/2}$. Improving the dependence on $\varepsilon$ for positive radii while retaining the dependence on $A_*$ would therefore require an analysis or an estimator that goes beyond this fixed-batch bias bound.
\end{remark}

\begin{remark}[Comparison with the bound for the same model]\label{rem:same-model}
Our criterion is the robust Markov CCE of \citet{shi2024sample} (see the discussion after Definition~\ref{def:cce}). Up to logarithmic factors, the ratio of our upper bound to theirs \cite[Theorem~4.1]{shi2024sample} is $A_*H^2D^3/(\varepsilon^2\prod_iA_i)$, so the two rates cross at $\prod_iA_i/A_*=H^2D^3/\varepsilon^2$. Their Theorem~4.2 gives a lower bound of order $SH^3DA_*/\varepsilon^2$ for the sampling scheme considered there: with fewer samples, every estimator fails with probability at least $1/8$ on some instance. This lower bound assumes $0<\rho_i\le1-c_0$ for a constant $c_0\in(0,1/4]$, $H>16\log2$, and the accuracy condition in their equation~(25). Our upper bound has the same powers of $S$ and $A_*$ as this benchmark. Establishing sharpness for our sampling model would require a lower bound that also covers adaptive selection of the queried rows.
\end{remark}

\begin{remark}[Computation]\label{rem:computation}
At the stages $h<H$, Algorithm~\ref{alg:backward} never enumerates the joint actions. A round of Algorithm~\ref{alg:local} costs $O(\sum_iA_i+mn\log n)$ arithmetic operations, and the constants $c_i$ are computed once per stage in $O(mS)$ operations. At stage $H$, an exact CCE can be found by solving the linear program~\eqref{glob:terminal}, whose size is polynomial in $\prod_iA_i$. Remark~\ref{rem:terminal-bandit} in Appendix~\ref{app:global} avoids this cost with a reward-only version of Algorithm~\ref{alg:local}, which keeps the gap bound $\varepsilon$ and still makes no transition query at stage $H$.
\end{remark}

\section{Proof overview}\label{sec:overview}

The proof of Theorem~\ref{thm:simple-global} has three steps, which correspond to Appendices~\ref{est:section}, \ref{sec:local}, and~\ref{app:global}. The first step bounds the statistical error of one plug-in backup. The second shows that a local run of Algorithm~\ref{alg:local} returns an approximate equilibrium of the local game together with accurate payoff estimates. The third propagates the local errors through the backward recursion.

\textbf{Step 1: one batch estimates a robust backup with a mean error that does not depend on $S$.}
Fix a row $p$, a continuation vector $u$, and a radius $\rho$, and let $\widehat p_n$ be the empirical distribution of $n$ independent samples from $p$. Proposition~\ref{est:bias} shows that
\begin{equation}\label{ov:bias}
 \bigl|\E\,\rob_\rho(\widehat p_n,u)-\rob_\rho(p,u)\bigr|
 \le\frac{\spn(u)}{2\sqrt n}.
\end{equation}
The proof starts from the standard dual form of the TV-constrained problem, which expresses the robust backup as an integral of clipped tail probabilities (Lemma~\ref{est:tail}):
\[
 \rob_\rho(p,u)=u_{\min}+\int_0^{\spn(u)}\bigl(p\{u-u_{\min}\one>\tau\}-\rho\bigr)_+\,d\tau .
\]
For each threshold $\tau$, the empirical tail probability is an average of $n$ Bernoulli variables, so its mean absolute error is at most $1/(2\sqrt n)$. Since the positive part is $1$-Lipschitz, integrating over $\tau$ gives~\eqref{ov:bias}. Two features of this bound matter for the rest of the proof. It involves only scalar tail probabilities, so it does not depend on the number of states, whereas estimating the row itself in TV distance would cost a factor that grows with $S$. It is also a deterministic bound on the mean of the estimate that holds at every row, so the algorithm can query only the row of the realized joint action, and no union bound over $\cA$ is needed.

\textbf{Step 2: the shared feedback certifies an approximate equilibrium of the local game.}
Fix a stage $h<H$ and a state $s$, and condition on the simulator responses, sampled actions, and internal randomness used before the run at $(h,s)$, so that the vectors $U_{i,h+1}$ are fixed. For the learned law $\pi_h(\cdot\mid s)$, define the payoff of obeying and the best payoff of a fixed action in the local game with continuation vectors $U_{i,h+1}$ by
\begin{align}
 C_{i,h}(s)
 &\coloneqq\E_{a\sim\pi_h(\cdot\mid s)}
    \bigl[r_{i,h}(s,a)+\rob_{\rho_i}(P_h(\cdot\mid s,a),U_{i,h+1})\bigr],
    \label{glob:obedience-backup}\\
 M_{i,h}(s)
 &\coloneqq\max_{b\in\cA_i}
   \E_{a_{-i}\sim\pi_{-i,h}(\cdot\mid s)}
    \bigl[r_{i,h}(s,(b,a_{-i}))
       +\rob_{\rho_i}(P_h(\cdot\mid s,(b,a_{-i})),U_{i,h+1})\bigr].
    \label{glob:deviation-backup}
\end{align}
These are the right-hand sides of the robust Bellman equations~\eqref{eq:obey} and~\eqref{eq:br}, with the learned vector $U_{i,h+1}$ in place of the true continuation values. Theorem~\ref{loc:certificate} and Corollary~\ref{loc:accuracy} show that, with the parameters in~\eqref{glob:parameters} and with probability at least $1-\delta/(S(H-1))$, simultaneously for all players,
\begin{equation}\label{ov:certificate}
 \bigl|U_{i,h}(s)-C_{i,h}(s)\bigr|\le\epsz
 \qquad\text{and}\qquad
 M_{i,h}(s)-U_{i,h}(s)\le\epsz .
\end{equation}
The scalar $U_{i,h}(s)$ is thus both an accurate estimate of the payoff of obeying and an approximate upper bound on the payoff of every fixed-action deviation.

The difficulty in proving~\eqref{ov:certificate} is that player~$i$ never observes the payoffs of the actions it did not play, and the payoff that it does observe is random through both the other players' actions and the batch. To obtain a bandit problem, the analysis draws, given the past, the opponents' actions $a_{-i,t}$ and a table of potential batches, one from each row, independently of $a_{i,t}$. The loss of each candidate action $b$ uses the table entry for $(b,a_{-i,t})$. This fixes the whole loss vector before $a_{i,t}$ is drawn, while the algorithm queries only the selected row. The conditional mean loss of $b$ is determined by the \emph{mean game}, whose payoff at a joint action is the expected plug-in payoff at its row; see~\eqref{loc:comparator-mean}. The high-probability analysis of EXP3-IX~\cite{neu2015} and the Azuma--Hoeffding inequality bound the regret of player~$i$ against these conditional means (Lemma~\ref{loc:ix}), which gives~\eqref{ov:certificate} for the mean game. The bias bound~\eqref{ov:bias} then transfers both inequalities to the local game at a cost of $\spn(U_{i,h+1})/(2\sqrt{n_h})\le\epsz/2$. Because the players are analyzed one at a time and combined by a union bound, the feedback of different players need not be independent.

\textbf{Step 3: the local errors add up along the backward recursion.}
By a union bound over the $S(H-1)$ local runs, the inequalities~\eqref{ov:certificate} hold at all stages $h<H$ and all states simultaneously with probability at least $1-\delta$. At stage $H$ they hold with zero error, because $\pi_H(\cdot\mid s)$ is an exact CCE of the reward game. The robust backup is monotone in the continuation vector and commutes with the addition of constants (Lemma~\ref{est:range}). Comparing~\eqref{glob:obedience-backup} and~\eqref{glob:deviation-backup} with the robust Bellman equations of Lemma~\ref{lem:bellman}, backward induction therefore gives (Lemma~\ref{glob:propagation})
\[
 U_{i,h}-V_{i,h}^{\pi}\le(H-h)\,\epsz\,\one
 \qquad\text{and}\qquad
 V_{i,h}^{\dagger,\pi_{-i}}-U_{i,h}\le(H-h)\,\epsz\,\one
\]
for every player and stage. Adding the two inequalities at $h=1$ bounds the robust CCE gap at every initial state by $2(H-1)\epsz\le\varepsilon$. The query count follows by multiplying the number of rounds by the batch size and summing over the stages and states.

\section{Conclusion}\label{sec:conclusion}
We studied robust Markov games in which every player has a TV ball around each nominal row, indexed by stage, state, and joint action. The robust backup is nonlinear in the row, so averaging nominal rows over the opponents' actions can change the payoff, and no estimate of the backup from a single successor is unbiased at every row. We showed that a batch of samples from the row of the realized joint action controls this nonlinearity in the mean, with an error that does not depend on the number of states. The resulting plug-in payoffs can then serve as shared bandit feedback for all players. Combined with backward induction, these two facts yield a robust Markov CCE from $\widetilde O(SA_*H^5D^4/\varepsilon^4)$ samples, a bound that depends on the action sets only through the largest individual one. The curse of multiagency can therefore be broken for $(s,a)$-rectangular TV uncertainty sets without changing their structure.

\textbf{Limitations and future work.}
A natural next question is whether the dependence on $1/\varepsilon$ can be made quadratic at positive radii while retaining the dependence on $A_*$. The lower bound of \citet[Theorem~4.2]{shi2024sample} provides a benchmark in the parameter range described in Remark~\ref{rem:same-model}, but a sharp rate for adaptive queries also requires a lower bound for that sampling model. Our analysis relies on the tail-integral form of the TV backup, and it would be interesting to obtain analogous mean-error bounds for other divergences, such as the Kullback--Leibler or the $\chi^2$ divergence. Finally, the generative model removes the exploration problem; extending the shared-batch idea to online interaction is left for future work.

\bibliographystyle{abbrvnat}
\bibliography{iclr2027_conference}

\clearpage
\appendix

\section*{Organization of the appendix and notation}
The appendix proves Theorem~\ref{thm:simple-global} in three parts, which follow the three steps of Section~\ref{sec:overview}. Appendix~\ref{est:section} studies the robust backup $\rob_\rho(p,u)$ and its plug-in estimate from one batch. Appendix~\ref{sec:local} analyzes Algorithm~\ref{alg:local} for a single local game. Appendix~\ref{app:global} combines the local guarantees along the backward recursion and proves the theorem. Appendix~\ref{app:bellman} proves the robust Bellman equations of Lemma~\ref{lem:bellman}. Table~\ref{tab:notation} collects the notation used across these parts.

\begin{table}[h]
\centering
\caption{Notation.}\label{tab:notation}
\smallskip
\begin{tabular}{@{}ll@{}}
\toprule
Symbol & Meaning\\
\midrule
$m$, $S$, $H$ & numbers of players, states, and stages\\
$A_i$, $A_*$ & number of actions of player~$i$; $A_*\coloneqq\max_iA_i$\\
$\rho_i$, $D_i$, $D$ & TV radius of player~$i$; $D_i\coloneqq\min\{H,1/\rho_i\}$; $D\coloneqq\max_iD_i$\\
$\rob_\rho(p,u)$ & robust backup $\min\{qu:q\in\Delta(\cS),\ \TV(q,p)\le\rho\}$\\
$u_{\min}$, $\spn(u)$ & minimum $\min_su(s)$ and span $\max_su(s)-u_{\min}$ of a vector $u$\\
$x_+$ & positive part $\max\{x,0\}$\\
$\widehat p_n$ & empirical distribution of a batch of $n$ successors\\
$d_{i,h}$, $d_h$ & span bounds~\eqref{glob:range-recursion}; $d_h\coloneqq\max_id_{i,h}$\\
$\epsz$, $\Lambda$, $T_h$, $n_h$ & local accuracy, confidence parameter, numbers of rounds, batch sizes~\eqref{glob:parameters}\\
$U_{i,h}$ & continuation value estimates computed by Algorithm~\ref{alg:backward}\\
$g_i$, $\bar g_i$, $\widehat g_{i,t}$ & payoff of the local game, its mean plug-in version~\eqref{loc:mean-game}, and the observed payoff\\
$\beta_i$ & bias $\max_a|\bar g_i(a)-g_i(a)|\le\spn(v_i)/(2\sqrt n)$ of the plug-in payoff~\eqref{loc:mean-game}\\
$C_i$, $M_i$ & payoff of obeying and best fixed-action payoff in a local game~\eqref{loc:exact-values}\\
$C_{i,h}$, $M_{i,h}$ & the same quantities at stage $h$, see~\eqref{glob:obedience-backup}--\eqref{glob:deviation-backup}\\
\bottomrule
\end{tabular}
\end{table}

\section{The robust backup and its fixed-batch estimate}\label{est:section}

This appendix establishes the properties of the robust backup on which the rest of the proof relies. Lemma~\ref{est:tail} gives two representations of $\rob_\rho(p,u)$, Lemma~\ref{est:range} derives its range and monotonicity properties, and Proposition~\ref{est:bias} bounds the mean error of the plug-in estimate $\rob_\rho(\widehat p_n,u)$. Throughout, $p\in\Delta(\cS)$ is a distribution, $\rho\in[0,1]$ is a radius, and $u\in\mathbb R^{\cS}$ is a vector. We center $u$ at its minimum,
\[
 y\coloneqq u-u_{\min}\one ,
\]
so that $y\ge0$, $\min_sy(s)=0$, and $\max_sy(s)=\spn(u)$. For a threshold $\tau\ge0$, we write $p\{y>\tau\}\coloneqq\sum_{s:\,y(s)>\tau}p(s)$ for the tail probability of $y$ under $p$.

\subsection{Two representations of robust backup}

The first representation below is the standard dual form of the robust backup, which reduces the minimization over distributions to a one-dimensional maximization over a clipping level $\alpha$. The second, which we derive from the first, expresses the backup through the tail probabilities of $y$. It is the form that we use to analyze the plug-in estimate, because each tail probability can be estimated separately.

\begin{lemma}[Clipping and tail representations]\label{est:tail}
For every $p\in\Delta(\cS)$, $\rho\in[0,1]$, and $u\in\mathbb R^{\cS}$,
\begin{align}
 \rob_\rho(p,u)
 &=u_{\min}+\max_{0\le\alpha\le\spn(u)}
       \Bigl\{\sum_{s\in\cS}p(s)\min\{y(s),\alpha\}-\rho\alpha\Bigr\}
 \label{est:clip-formula}\\
 &=u_{\min}+\int_0^{\spn(u)}\bigl(p\{y>\tau\}-\rho\bigr)_+\,d\tau .
 \label{est:tail-formula}
\end{align}
In particular, $\rob_0(p,u)=pu$ and $\rob_1(p,u)=u_{\min}$.
\end{lemma}

\begin{proof}
Write $L=\spn(u)$ and $G(\tau)=p\{y>\tau\}$. Centering the vector separates the part of the payoff that nature can change. Every feasible $q$ is a
probability distribution, so $qu=u_{\min}+qy$. It therefore suffices to compute the worst-case expectation of
$y$, whose values lie in $[0,L]$. If $L=0$, then $y=0$ and both right-hand sides reduce to $u_{\min}$, which is $qu$ for every row. Suppose henceforth that $L>0$.

\emph{Apply the TV clipping identity.}
For $0<\rho\le1$, \citet[Lemma~B.1 and eq.~(23)]{shi2024sample} express the minimum expectation
over a TV ball as a maximization over clipping levels. Their result takes a probability row, a ball
defined by half the $\ell_1$ distance on the full simplex, and a nonnegative continuation vector.
Our nominal row $p$, radius $\rho$, and centered vector $y$ satisfy these hypotheses, including
when some entries of $p$ are zero. The clipping levels range from the minimum of $y$ to its maximum,
which here gives $[0,L]$. For any such $\alpha$, the clipped vector $\min\{y,\alpha\}$ has minimum zero,
so the penalty in the cited formula reduces to $\rho\alpha$. Adding back $u_{\min}$ gives
\eqref{est:clip-formula}. At $\rho=0$ the only feasible row is $p$ and the maximizing level
$\alpha=L$ gives $py$, so the same formula holds there as well.

\emph{Pass from the clipping level to tail probabilities.}
For any $\alpha\in[0,L]$, the amount of $y(s)$ below level $\alpha$ is
$\min\{y(s),\alpha\}=\int_0^\alpha\ind{y(s)>\tau}\,d\tau$. Summing over the finite state space with
weights $p(s)$ gives
\begin{equation}\label{est:objective-integral}
 \sum_s p(s)\min\{y(s),\alpha\}-\rho\alpha
      =\int_0^\alpha\bigl(G(\tau)-\rho\bigr)\,d\tau .
\end{equation}
As the level increases, the integrand is first positive, then possibly zero, and then nonpositive, because
the sets $\{s:y(s)>\tau\}$ shrink as $\tau$ increases. More precisely, take $\alpha^\star$ to be the supremum of the
levels where $G(\tau)>\rho$, or zero if there are none. Integrating up to $\alpha^\star$ keeps the positive
part and discards the negative part. The value of the integrand at a single endpoint does not affect the integral.
It follows from~\eqref{est:objective-integral} that the maximum in~\eqref{est:clip-formula} is
$\int_0^L(G(\tau)-\rho)_+\,d\tau$. This proves~\eqref{est:tail-formula}. At $\rho=1$ its integrand
vanishes everywhere, giving the endpoint value $u_{\min}$.

\emph{A minimizing row for all thresholds.}
The tail formula also identifies which probability mass nature can move away from high-value states.
For $\tau\in[0,L]$, set $B_\tau=\{s:y(s)>\tau\}$. The total mass by which a probability distribution $q$
can reduce $p$ on any subset is at most $\sum_s(p(s)-q(s))_+=\tfrac12\|p-q\|_1$. The equality uses
$\sum_s(p(s)-q(s))=0$. Consequently every feasible $q$ satisfies
$q(B_\tau)\ge(p(B_\tau)-\rho)_+$. We now choose one feasible row that attains all these lower bounds.

Choose a state $s_0$ with $y(s_0)=0$ and let $\lambda=\min\{\rho,p(B_0)\}$. Remove a total mass $\lambda$
from the states with positive $y$, starting at the largest value of $y$ and proceeding in decreasing order.
At each state remove at most its nominal mass, taking only the remaining amount if the budget is exhausted
there. Add all the removed mass to $s_0$. The resulting row $q^\star$ is nonnegative and sums to one.
Its TV distance from $p$ is $\lambda$, since the removed and added masses have disjoint supports.
The construction remains feasible if $p(s_0)=0$, because the uncertainty set is a ball in the full simplex.

For any fixed $\tau$, all states in $B_\tau$ are visited before any state with $0<y(s)\le\tau$.
Thus the amount removed from $B_\tau$ is exactly $\min\{\lambda,p(B_\tau)\}$, irrespective of how ties
were ordered. Since $s_0\notin B_\tau$ and $p(B_\tau)\le p(B_0)$, this gives
\[
 q^\star(B_\tau)=p(B_\tau)-\min\{\lambda,p(B_\tau)\}
               =(p(B_\tau)-\rho)_+ .
\]
Finally, $y(s)=\int_0^L\ind{s\in B_\tau}\,d\tau$, so integrating the last equality gives
$q^\star y=\int_0^L(G(\tau)-\rho)_+\,d\tau$. Thus the same feasible row attains the robust value
and realizes the tail reduction at every threshold.
\end{proof}

The tail representation has a simple interpretation. A minimizing distribution removes mass from the states with the largest values of $u$, up to a total mass of $\rho$, and moves it to a state at which $u$ attains its minimum. This lowers every tail probability $p\{y>\tau\}$ by $\rho$, or to zero if the tail is lighter than $\rho$. Because the ball is taken in the full simplex, the receiving state may have zero nominal probability, which is why $u_{\min}$ is the minimum over all of $\cS$ and not only over the support of $p$. The clipping representation~\eqref{est:clip-formula} also shows that $p\mapsto\rob_\rho(p,u)$ is a maximum of affine functions of $p$ and is therefore convex; this is the convexity invoked in the introduction.

\subsection{Range and monotonicity}

The next lemma collects the elementary properties of the robust backup that are used in Appendices~\ref{sec:local} and~\ref{app:global}. Part~(a) bounds the range of the payoffs in the local game, and parts~(b) and~(c) are the properties that allow errors in the continuation values to be propagated through the backward recursion.

\begin{lemma}[Range, monotonicity, and constant shifts]\label{est:range}
Let $p\in\Delta(\cS)$, $\rho\in[0,1]$, and $u,v\in\mathbb R^{\cS}$.
\begin{enumerate}[label=\textup{(\alph*)}]
\item \textup{(Range.)} $u_{\min}\le\rob_\rho(p,u)\le u_{\min}+(1-\rho)\,\spn(u)$. Consequently, for every $r\in[0,1]$,
\begin{equation}\label{est:range-bound}
 0\le r+\rob_\rho(p,u)-u_{\min}\le1+(1-\rho)\,\spn(u).
\end{equation}
\item \textup{(Monotonicity.)} If $u\le v$, then $\rob_\rho(p,u)\le\rob_\rho(p,v)$.
\item \textup{(Constant shifts.)} For every $c\in\mathbb R$, $\rob_\rho(p,u+c\one)=\rob_\rho(p,u)+c$.
\end{enumerate}
In particular, if $u\le v+c\one$ for some $c\in\mathbb R$, then
\begin{equation}\label{glob:one-sided-lipschitz}
 \rob_\rho(p,u)\le\rob_\rho(p,v)+c .
\end{equation}
All statements hold for every distribution $p$ on $\cS$, and in particular for empirical distributions.
\end{lemma}

\begin{proof}
The range will bound every realized payoff in the bandit update. Since $G(\tau)=p\{y>\tau\}$ is a probability,
$0\le(G(\tau)-\rho)_+\le1-\rho$. The integral in~\eqref{est:tail-formula} has length $\spn(u)$, so
\[
 0\le\rob_\rho(p,u)-u_{\min}\le(1-\rho)\spn(u).
\]
Adding $r\in[0,1]$ gives~\eqref{est:range-bound}. Both endpoints of this interval are independent of $p$,
which allows us to use it for an empirical row as well.

The remaining properties concern the continuation vector, with the row set held fixed. If $u\le v$,
then $qu\le qv$ for each feasible $q$, since its entries are nonnegative. Minimizing over the same set
proves monotonicity. Also $q(u+c\one)=qu+c$ because $q\one=1$. The constant is the same for all feasible
$q$, so it passes through the minimum. Applying monotonicity first to $u\le v+c\one$ and then taking the
constant outside the second backup gives~\eqref{glob:one-sided-lipschitz}.
\end{proof}

\subsection{The fixed-batch estimate}

We now turn to estimation. Let $X_1,\dots,X_n$ be $n\ge1$ independent samples from $p$, and let $\widehat p_n\coloneqq n^{-1}\sum_{j=1}^n\delta_{X_j}$ be their empirical distribution. The plug-in estimate of $\rob_\rho(p,u)$ is $\rob_\rho(\widehat p_n,u)$, the robust backup of the empirical distribution. By Lemma~\ref{est:tail} with $\widehat p_n$ in place of $p$, it can be computed from the batch as
\[
 \rob_\rho(\widehat p_n,u)
 =u_{\min}+\max_{0\le\alpha\le\spn(u)}
       \Bigl\{\frac1n\sum_{j=1}^n\min\{y(X_j),\alpha\}-\rho\alpha\Bigr\},
\]
which requires no knowledge of $p$; see the discussion after~\eqref{alg:plugin}. The estimate is biased in general, because the backup is a nonlinear function of the distribution. The following proposition shows that its mean error is nevertheless of order $n^{-1/2}$, with a constant that involves only the span of $u$.

\begin{proposition}[Mean error of the plug-in backup]\label{est:bias}
For every $p\in\Delta(\cS)$, $u\in\mathbb R^{\cS}$, $\rho\in[0,1]$, and $n\ge1$,
\begin{equation}\label{est:mean-error}
 \E\bigl|\rob_\rho(\widehat p_n,u)-\rob_\rho(p,u)\bigr|
 \le\frac{\spn(u)}{2\sqrt n},
\end{equation}
and consequently
\begin{equation}\label{est:bias-bound}
 \bigl|\E\,\rob_\rho(\widehat p_n,u)-\rob_\rho(p,u)\bigr|
 \le\frac{\spn(u)}{2\sqrt n}.
\end{equation}
\end{proposition}

\begin{proof}
The proof uses the tail representation twice, once for the absolute error of the estimate on a realized
batch and once for the sign of its average error. Let $L=\spn(u)$ and $y=u-u_{\min}\one$ as before.
At threshold $\tau$, write
\[
 \widehat G_n(\tau)=\frac1n\sum_{j=1}^n\ind{y(X_j)>\tau},\qquad
 G(\tau)=\Prob\{y(X_1)>\tau\}.
\]
For each \emph{fixed} threshold the indicators are independent Bernoulli variables with common mean
$G(\tau)$, because the $X_j$ are independent draws from $p$. Thus $\E\widehat G_n(\tau)=G(\tau)$ and
\[
 \E\bigl|\widehat G_n(\tau)-G(\tau)\bigr|
 \le \sqrt{\E\bigl(\widehat G_n(\tau)-G(\tau)\bigr)^2}
 = \sqrt{\frac{G(\tau)(1-G(\tau))}{n}}\le \frac1{2\sqrt n}.
\]
The first inequality is Cauchy--Schwarz and the last uses the maximum variance of a Bernoulli variable.
The empirical tails at different thresholds share samples. The argument only integrates their separate
error bounds.

\emph{Absolute error.}
For each realized batch $\widehat p_n$ is a probability vector on $\cS$. Lemma~\ref{est:tail} therefore
applies to it with the same $y$, $L$, and $\rho$ as for $p$. Subtracting the two integrals, taking absolute
values and using $|(z-\rho)_+-(w-\rho)_+|\le|z-w|$ gives
\begin{equation}\label{est:pathwise}
 \bigl|\rob_\rho(\widehat p_n,u)-\rob_\rho(p,u)\bigr|
 \le\int_0^L|\widehat G_n(\tau)-G(\tau)|\,d\tau .
\end{equation}
The integrand is a nonnegative measurable step function. Tonelli's theorem permits us to take its
expectation inside the integral. Integrating the Bernoulli bound over an interval of length $L$ proves
\eqref{est:mean-error}.

\emph{Signed bias.}
Write $X=\widehat G_n(\tau)$ and $x=G(\tau)$ at a fixed threshold. The function $z\mapsto(z-\rho)_+$
is convex, and $X\in[0,1]$ has mean $x$. Jensen's inequality therefore gives
$\E(X-\rho)_+\ge(x-\rho)_+$. To bound the size of this difference, use $z_+=(z+|z|)/2$.
The linear terms cancel because $\E X=x$, and the triangle inequality gives
\[
 \E(X-\rho)_+-(x-\rho)_+
   =\tfrac12\bigl(\E|X-\rho|-|x-\rho|\bigr)
   \le \tfrac12\E|X-x|
   \le \frac1{4\sqrt n}.
\]
Both clipped tails lie in $[0,1]$ and $L$ is finite, so their difference is integrable on the product of
the sample space and $[0,L]$. Fubini's theorem now identifies the signed bias with the integral of these
threshold biases. Integrating their lower and upper bounds yields the sharper signed estimate
\[
 0\le\E\,\rob_\rho(\widehat p_n,u)-\rob_\rho(p,u)\le\frac{L}{4\sqrt n}.
\]
In particular its absolute value is at most $L/(2\sqrt n)$ as claimed in~\eqref{est:bias-bound}.
The sign will also let us bound the payoff of a deviation directly in the local game.

If $L=0$, both integrals vanish on every sample path. At $\rho=0$ the empirical backup is
$n^{-1}\sum_j u(X_j)$ and its expectation is $pu$. At $\rho=1$ the empirical and true backups both equal
the minimum of $u$ over $\cS$. Thus the signed bias vanishes at both endpoint radii.
\end{proof}

\begin{remark}[Dimension-free error and sharpness of the plug-in bias]\label{est:dimension-free}
For each realized batch, the right-hand side of~\eqref{est:pathwise} is the one-dimensional
Wasserstein-$1$ distance between the empirical and nominal laws of $y(X)$. To see this, let
$Y$ have the nominal law of $y(X)$ and $\widehat Y$ the empirical law, under any coupling. For each outcome, $|Y-\widehat Y|$
is the length of the set of thresholds at which their tail indicators differ. For any threshold,
the probability of this disagreement is at least $|\widehat G_n(\tau)-G(\tau)|$.
If both variables are drawn by applying their respective quantile functions to the same
uniform random variable, the two tail events are nested intervals of that variable, so equality
holds at every threshold. Integrating over $[0,\spn(u)]$ gives the claimed minimum expected
distance. This explains why the bound depends only on scalar tail probabilities.

The $n^{-1/2}$ dependence in~\eqref{est:bias-bound} occurs even for $u=(0,1)$, $p=(1/2,1/2)$, and
$\rho=1/2$. Let $K$ count the successors with value one. Then $K\sim\mathrm{Bin}(n,1/2)$, and
Lemma~\ref{est:tail} gives true backup zero and empirical backup $(K/n-1/2)_+$. Put
$Z_n=(2K-n)/\sqrt n$. The distribution of $Z_n$ is symmetric because $K$ and $n-K$ have the same law.
Consequently
\[
 \E(K/n-1/2)_+=\frac{\E(Z_n)_+}{2\sqrt n}
              =\frac{\E|Z_n|}{4\sqrt n}.
\]
To evaluate the last expectation asymptotically, write $Z_n$ as $1/\sqrt n$ times a sum of $n$
independent, identically distributed signs with mean zero and variance one. The central limit theorem
gives $Z_n\Rightarrow Z$ for $Z\sim N(0,1)$. To pass to the unbounded function $|z|$, first truncate it
at $M>0$. The truncated function is bounded and continuous, so its expectations converge. The discarded
tail is uniformly small because $\E Z_n^2=1$ and
$\E[|Z_n|\ind{|Z_n|>M}]\le1/M$. The same estimate holds for $Z$. Letting $M$ tend to infinity gives
$\E|Z_n|\to\E|Z|=\sqrt{2/\pi}$. Hence the signed bias is asymptotic to $1/\sqrt{8\pi n}$. Relative to the upper bound in~\eqref{est:bias-bound}, the ratio tends to
$1/\sqrt{2\pi}$.
\end{remark}

\begin{remark}[One batch, several vectors]\label{est:shared-batch}
In a round of Algorithm~\ref{alg:local}, condition on the past and the selected joint action. The row, continuation
vectors, and radii are then fixed, and the batch consists of independent draws from that row.
Proposition~\ref{est:bias} therefore applies to each player's estimate conditional on the same information. Reusing
successors across players preserves each of these conditional bounds.
\end{remark}

\section{The local game: analysis of Algorithm~\ref{alg:local}}
\label{sec:local}

This appendix analyzes one run of Algorithm~\ref{alg:local}. Its main result, Theorem~\ref{loc:certificate}, shows that the output $(\mu,U_1,\dots,U_m)$ certifies an approximate equilibrium of the local game: each scalar $U_i$ is close to the payoff of player~$i$ under $\mu$, and no fixed action of player~$i$ earns much more than $U_i$ against the other players' marginal of $\mu$.

\subsection{Setting and statement}

Throughout this appendix the stage and the state are fixed and suppressed. The inputs of Algorithm~\ref{alg:local} are deterministic: a row $p_a\in\Delta(\cS)$ for every joint action $a\in\cA$, and, for every player~$i$, a reward function $r_i:\cA\to[0,1]$, a radius $\rho_i\in[0,1]$, and a continuation vector $v_i\in\mathbb R^{\cS}$. Remark~\ref{loc:conditional} extends the analysis to inputs that depend on earlier data. Recall from~\eqref{alg:local-game} that the local game has the payoffs
\[
 g_i(a)=r_i(a)+\rob_{\rho_i}(p_a,v_i),
\]
and recall that $c_i=\min_sv_i(s)$. We assume that the range parameter satisfies
\begin{equation}\label{loc:range-condition}
 R\ge\max_{i\in[m]}\bigl\{1+(1-\rho_i)\,\spn(v_i)\bigr\}.
\end{equation}
By the range bound~\eqref{est:range-bound}, this condition guarantees that
\begin{equation}\label{loc:centered-range}
 0\le r_i(a)+\rob_{\rho_i}(p,v_i)-c_i\le R
 \qquad\text{for all }i\in[m],\ a\in\cA,\text{ and }p\in\Delta(\cS).
\end{equation}
In particular, every loss $\ell_{i,t}$ computed in line~8 of Algorithm~\ref{alg:local} lies in $[0,1]$, since it is obtained from~\eqref{loc:centered-range} with the empirical distribution $p=\widehat p_t$.

\textbf{The mean game.}
The plug-in payoff observed at a joint action $a$ is an unbiased estimate of its expectation over the batch, not of $g_i(a)$. For a joint action $a$, let $\widehat p_{a,n}$ denote the empirical distribution of $n$ independent samples from $p_a$, and define the payoffs of the \emph{mean game} and their largest deviation from the local game by
\begin{equation}\label{loc:mean-game}
 \bar g_i(a)\coloneqq r_i(a)+\E\bigl[\rob_{\rho_i}(\widehat p_{a,n},v_i)\bigr],
 \qquad
 \beta_i\coloneqq\max_{a\in\cA}\bigl|\bar g_i(a)-g_i(a)\bigr| .
\end{equation}
The bias bound~\eqref{est:bias-bound} of Proposition~\ref{est:bias}, applied with $p=p_a$ and $u=v_i$ for every joint action $a$, gives
\begin{equation}\label{loc:plugin-bias}
 \beta_i\le\frac{\spn(v_i)}{2\sqrt n}\qquad\text{for every }i\in[m].
\end{equation}
Moreover, $\beta_i=0$ if $\rho_i=0$, because $\rob_0(\widehat p_{a,n},v_i)=\widehat p_{a,n}v_i$ by Lemma~\ref{est:tail} and $\E[\widehat p_{a,n}v_i]=p_av_i$. Since $\beta_i$ compares two deterministic payoffs, the bound~\eqref{loc:plugin-bias} applies at every row, including rows that are never sampled.

\textbf{Parameters and output.}
The algorithm is run for $T\ge1$ rounds with batches of size $n\ge1$ and with the confidence parameter
\[
 \Lambda\coloneqq\log\frac{8mA_*}{\delta},\qquad\text{so that}\qquad
 \gamma_i=\sqrt{\frac{\Lambda}{A_iT}},
\]
for some $\delta\in(0,1)$. Its output is the joint distribution $\mu=T^{-1}\sum_{t=1}^Tx_t$, where $x_t=\bigotimes_ix_{i,t}$, together with the averages $U_i=T^{-1}\sum_{t=1}^T\widehat g_{i,t}$ of the observed payoffs. Let $\mu_{-i}$ denote the marginal of $\mu$ on the actions of the players other than~$i$. We compare $U_i$ with two exact quantities of the local game, the payoff of player~$i$ under $\mu$ and its best payoff from a fixed action against $\mu_{-i}$:
\begin{equation}\label{loc:exact-values}
 C_i\coloneqq\E_{a\sim\mu}\,g_i(a),\qquad
 M_i\coloneqq\max_{b\in\cA_i}\E_{a_{-i}\sim\mu_{-i}}\,g_i(b,a_{-i}).
\end{equation}
Both quantities are random, because $\mu$ depends on the data.

\begin{theorem}[Local certificate from shared batches]
\label{loc:certificate}
Assume~\eqref{loc:range-condition} and run Algorithm~\ref{alg:local} with $T\ge1$ rounds, batch size $n\ge1$, and $\Lambda=\log(8mA_*/\delta)$, where $\delta\in(0,1)$. The run makes exactly $nT$ queries. With probability at least $1-\delta$, the following two bounds hold simultaneously for every player $i\in[m]$:
\begin{align}
 |C_i-U_i|
 &\le\beta_i+R\sqrt{\frac{2\Lambda}{T}},
       \label{loc:obedience-bound}\\
 M_i-U_i
 &\le\beta_i+R\left(5\sqrt{\frac{A_i\Lambda}{T}}
                       +\frac{\Lambda}{T}\right),
       \label{loc:deviation-bound}
\end{align}
where the bias $\beta_i$ defined in~\eqref{loc:mean-game} satisfies $\beta_i\le\spn(v_i)/(2\sqrt n)$, and $\beta_i=0$ if $\rho_i=0$. Moreover, on every run, $c_i\le U_i\le c_i+1+(1-\rho_i)\,\spn(v_i)$ for every player~$i$.
\end{theorem}

In both bounds, the first term is the bias of the plug-in backup, which is controlled by the batch size $n$, and the second term is a learning error, which is controlled by the number of rounds $T$. Adding~\eqref{loc:deviation-bound} to~\eqref{loc:obedience-bound} bounds $M_i-C_i$, which shows that $\mu$ is an approximate CCE of the local game. The backward recursion needs slightly more than this, namely the two bounds separately, because $U_i$ is the quantity that is passed on to the preceding stage. The next corollary records a choice of $T$ and $n$ that makes both errors at most a prescribed $\epsz$.

\begin{corollary}[Parameters for a prescribed accuracy]\label{loc:accuracy}
In the setting of Theorem~\ref{loc:certificate}, let $0<\epsz\le R$ and suppose that
\begin{equation}\label{loc:accuracy-parameters}
 n\ge\frac{\max_i\spn(v_i)^2}{\epsz^2}
 \qquad\text{and}\qquad
 T\ge\frac{256\,A_*R^2\Lambda}{\epsz^2}.
\end{equation}
Then, on the event of Theorem~\ref{loc:certificate}, $|C_i-U_i|\le\epsz$ and $M_i-U_i\le\epsz$ for every player~$i$, and consequently $M_i-C_i\le2\epsz$.
\end{corollary}

\begin{proof}
Work on the event in Theorem~\ref{loc:certificate}. Its two error bounds involve three terms of different
origins. The batch size controls the bias, while the round count controls the fluctuation under obedience
and the comparison with each fixed action.

First, $\spn(v_i)\le\epsz\sqrt n$ by the condition on $n$, so~\eqref{loc:plugin-bias} gives
$\beta_i\le\epsz/2$. For the remaining terms, the lower bound on $T$ is equivalent to
$\Lambda/T\le\epsz^2/(256A_*R^2)$. Substituting this inequality gives
\begin{align*}
 R\sqrt{2\Lambda/T}&\le\frac{\sqrt{2}}{16\sqrt{A_*}}\epsz\le\frac{\epsz}{8},\\
 5R\sqrt{A_i\Lambda/T}&\le\frac{5}{16}\sqrt{\frac{A_i}{A_*}}\,\epsz\le\frac{5\epsz}{16},\\
 R\Lambda/T&\le\frac{\epsz^2}{256 A_*R}\le\frac{\epsz}{256}.
\end{align*}
Here $1\le A_i\le A_*$, and the last inequality also uses $\epsz\le R$.
The obedience error in~\eqref{loc:obedience-bound} is therefore at most $\epsz/2+\epsz/8$.
The deviation error in~\eqref{loc:deviation-bound} is at most $\epsz/2+5\epsz/16+\epsz/256=209\epsz/256$. Each is at most
$\epsz$. Finally,
$M_i-C_i=(M_i-U_i)+(U_i-C_i)\le2\epsz$ for every player on the same event.
\end{proof}

With the smallest integers that satisfy~\eqref{loc:accuracy-parameters}, the run makes $nT=\widetilde O(A_*R^2\max_i\spn(v_i)^2/\epsz^4)$ queries when $\max_i\spn(v_i)\ge\epsz$. Only the largest individual action count $A_*$ enters polynomially, because the bandit updates are performed per player, and the number of players enters only through $\Lambda$, because all players reuse the same batch.

\subsection{Roadmap of the proof of Theorem~\ref{loc:certificate}}

The quantity $U_i$ is an average of observed payoffs, whereas $C_i$ and $M_i$ are expectations in the exact local game. The proof passes from one to the other through the mean game~\eqref{loc:mean-game}, in three steps.
\begin{enumerate}[label=\textup{(\arabic*)}]
\item \emph{Bias.} The mean game is uniformly close to the local game: $|\bar g_i(a)-g_i(a)|\le\beta_i$ for every joint action $a$, by the definition~\eqref{loc:mean-game}, and $\beta_i\le\spn(v_i)/(2\sqrt n)$ by~\eqref{loc:plugin-bias}.
\item \emph{Regret.} From the viewpoint of player~$i$, the rounds of Algorithm~\ref{alg:local} form a bandit problem. A high-probability regret bound for EXP3-IX shows that the average observed payoff $U_i$ is nearly as large as the payoff of every fixed action in the mean game against the other players' average play. This gives the deviation bound.
\item \emph{Concentration.} The average observed payoff $U_i$ is close to the average of its conditional expectations given the past rounds, and that average is the payoff of $\mu$ in the mean game. This gives the obedience bound.
\end{enumerate}
The main obstacle lies in step~(2). Player~$i$ observes only the payoff of the joint action that was actually played, and this payoff is random through both the other players' actions and the batch. A regret bound against the \emph{expected} payoff of each fixed action therefore does not follow directly from a bound for a fixed loss sequence. To overcome this, the analysis draws, given the past, a table of potential successor batches, one for each joint action, independently of the players' action draws. The opponents' actions and the table determine all counterfactual losses of player~$i$, they are jointly independent of its current action conditionally on the past, and the entry selected by the played action gives its observed loss. Lemma~\ref{loc:ix} states a regret bound for exactly this situation, and Section~\ref{sec:proof-certificate} verifies that Algorithm~\ref{alg:local} fits it.

\subsection{EXP3-IX with conditionally independent losses}\label{sec:ix}

We consider a single learner with $A\ge1$ actions over $T\ge1$ rounds. Let $\Lambda>0$ and $\gamma\coloneqq\sqrt{\Lambda/(AT)}$, and let $(\cF_t)_{t=0}^T$ be a filtration. In round $t$, the learner holds the exponential-weights distribution
\[
 x_t(b)\coloneqq\frac{w_t(b)}{\sum_{b'=1}^Aw_t(b')},\qquad w_1(b)\coloneqq1,
\]
the environment generates a loss vector $z_t\in[0,1]^A$, and the learner draws an action $I_t\in[A]$. The learner observes only the loss $z_t(I_t)$ of the action it played. It forms the implicit-exploration estimates and updates the weights by
\begin{equation}\label{loc:ix-update}
 \widehat z_t(b)\coloneqq\frac{z_t(b)\,\ind{I_t=b}}{x_t(b)+\gamma},
 \qquad
 w_{t+1}(b)\coloneqq w_t(b)\exp\{-\gamma\,\widehat z_t(b)\},
 \qquad b\in[A].
\end{equation}
This is the update in line~9 of Algorithm~\ref{alg:local} for one player, with the same value $\gamma$ for the learning rate and for the implicit-exploration parameter. We make the following assumption on the interaction.
\begin{equation}\label{loc:ix-assumption}
\parbox{0.88\linewidth}{For every $t\in[T]$, the pair $(z_t,I_t)$ is $\cF_t$-measurable, and conditionally on $\cF_{t-1}$ the action $I_t$ has distribution $x_t$ and is independent of the loss vector $z_t$.}
\end{equation}
Under this assumption, $x_t$ is $\cF_{t-1}$-measurable, because it is a function of the losses observed and the actions played before round $t$. The loss vector $z_t$ may depend on the past in an arbitrary way and may involve fresh randomness; what matters is that this randomness is independent of the learner's current draw. Define the conditional mean loss of each action and of the learner's strategy by
\[
 L_t(b)\coloneqq\E[z_t(b)\mid\cF_{t-1}],
 \qquad
 \bar L_t\coloneqq\sum_{b=1}^Ax_t(b)\,L_t(b).
\]

\begin{lemma}[EXP3-IX against conditional mean losses]
\label{loc:ix}
Assume~\eqref{loc:ix-assumption} and let $\Lambda\ge\log A$. Then there is an event of probability at least $1-(2A+3)e^{-\Lambda}$ on which, simultaneously for every action $b\in[A]$,
\begin{align}
 \sum_{t=1}^Tz_t(I_t)-\sum_{t=1}^TL_t(b)
 &\le5\sqrt{AT\Lambda}+\Lambda,
       \label{loc:ix-deviation}\\
 \Bigl|\sum_{t=1}^Tz_t(I_t)-\sum_{t=1}^T\bar L_t\Bigr|
 &\le\sqrt{2T\Lambda}.
       \label{loc:ix-obedience}
\end{align}
\end{lemma}

\begin{proof}
Write $Y_t=z_t(I_t)$ for the bounded loss that the learner observes. To prove the two inequalities,
we first compare $Y_t$ with an action's estimated loss on each sample path. We then derive a
conditional exponential bound to replace that estimate by $L_t(b)$. For the learner's own strategy,
the relevant comparison is with $\bar L_t$ and follows from concentration of a martingale.

The connection to the random loss vector in the statement is as follows. All $x_t(b)$ are positive
because the weights start positive and are multiplied by positive factors. Given $I_t=b$ we have
$Y_t=z_t(b)$. Conditional independence in~\eqref{loc:ix-assumption} therefore yields
\[
 \E[Y_t\mid\cF_{t-1},I_t=b]
 =\E[z_t(b)\mid\cF_{t-1},I_t=b]
 =\E[z_t(b)\mid\cF_{t-1}]=L_t(b).
\]
Averaging over $b$ gives $\E[Y_t\mid\cF_{t-1}]=\bar L_t$. Also,
$Y_t\ind{I_t=b}=z_t(b)\ind{I_t=b}$ on every sample path, so we can use the observed scalar in
the estimator~\eqref{loc:ix-update}.

\emph{Step 1. Compare the incurred loss with the estimates on the realized sample path.}
Write $\langle x,z\rangle=\sum_jx(j)z(j)$.
The deterministic exponential-weights inequality in \citet[eq.~(12.1)]{lattimore2020bandit} bounds the
cumulative estimated loss of a mixture relative to a fixed action. With uniform initial weights, exponential
updates of rate $\gamma>0$, and estimates satisfying $0\le\widehat z_t(j)\le1/x_t(j)$, it gives
\begin{equation}\label{loc:potential}
 \sum_{t=1}^T\langle x_t,\widehat z_t\rangle-\sum_{t=1}^T\widehat z_t(b)
 \le \frac{\log A}{\gamma}+
       \frac{\gamma}{2}\sum_{t=1}^T\sum_{j=1}^A\widehat z_t(j)
 \qquad (b\in[A]).
\end{equation}
The initialization and update are those of~\eqref{loc:ix-update}, and the weights are positive as shown above. An unplayed action has estimate zero, and the played
action has estimate $Y_t/(x_t(I_t)+\gamma)\le1/x_t(I_t)$ because $Y_t\in[0,1]$.
The inequality holds for each sequence of such estimates, so we may apply it before taking any
expectations over the random feedback.

Since only the played coordinate can be nonzero, we can express the loss the algorithm actually observes as
\[
 \langle x_t,\widehat z_t\rangle
 =\frac{x_t(I_t)Y_t}{x_t(I_t)+\gamma}
 =Y_t-\gamma\sum_{j=1}^A\widehat z_t(j).
\]
Substituting this relation in~\eqref{loc:potential} proves the pathwise inequality
\begin{equation}\label{loc:pathwise}
 \sum_{t=1}^T\bigl(Y_t-\widehat z_t(b)\bigr)
 \le \frac{\log A}{\gamma}
        +\frac{3\gamma}{2}\sum_{t=1}^T\sum_{j=1}^A\widehat z_t(j).
\end{equation}
We must control the estimate of action $b$ on the left and the sum of the estimates on the right.

\emph{Step 2. Control estimated losses relative to their conditional means.}
Even after $I_t=j$ is fixed, the observed loss can be random. Using the conditional mean identified
above and the sampling probability $x_t(j)$ gives
\[
 \E[\widehat z_t(j)\mid\cF_{t-1}]
 =\frac{x_t(j)}{x_t(j)+\gamma}\,L_t(j).
\]
The denominator makes this mean at most $L_t(j)$. A bound on the mean alone does not control the realized
sum in~\eqref{loc:pathwise}. For bandit losses, \citet[Lemma~1]{neu2015} bounds weighted sums of
IX estimates minus the underlying losses using nonnegative predictable weights at most $2\gamma$
for a constant exploration parameter $\gamma>0$. For a fixed set $J\subseteq[A]$ the weights
$2\gamma\ind{j\in J}$ satisfy this requirement. Our comparator $L_t(j)$ is the conditional mean of
a random observation, so we establish the corresponding one-round exponential bound directly.
We claim that
\begin{equation}\label{loc:ix-subset}
 \Prob\left\{\sum_{t=1}^T\sum_{j\in J}
       \bigl(\widehat z_t(j)-L_t(j)\bigr)>
             \frac{\Lambda}{2\gamma}\right\}\le e^{-\Lambda}.
\end{equation}

To see the one-round estimate behind this claim, put $p=x_t(I_t)>0$ and $u=2\gamma Y_t/p$.
The bound $Y_t\le1$ and the inequality $\log(1+u)\ge u/(1+u/2)$ give
\[
 \frac{2\gamma Y_t}{p+\gamma}
 \le \frac{2\gamma Y_t}{p+\gamma Y_t}
 =\frac{u}{1+u/2}\le\log(1+u).
\]
For the logarithmic inequality, $\log(1+u)$ is the integral of the convex function $z\mapsto1/(1+z)$
over $[0,u]$. Its average for $u>0$ is at least its value at the midpoint, and equality holds at $u=0$.
If the selected action lies in $J$, exponentiating the display bounds the exponential of its estimate by
$1+2\gamma Y_t/p$. If $I_t\notin J$, the exponential of the sum over $J$ equals one. Hence
\[
 \exp\!\left(2\gamma\sum_{j\in J}\widehat z_t(j)\right)
 \le 1+2\gamma\,\frac{Y_t\ind{I_t\in J}}{x_t(I_t)} .
\]
The expectation of the fraction conditional on $\cF_{t-1}$ is
$\sum_{j\in J}x_t(j)L_t(j)/x_t(j)=\sum_{j\in J}L_t(j)$.
This cancellation uses the definition of the conditional mean given the selected action, so it allows the
distribution of $Y_t$ to depend on that action. Using $1+v\le e^v$ and taking the predictable means outside
the conditional expectation gives
\begin{equation}\label{loc:one-round-moment}
 \E\!\left[\exp\!\left(2\gamma\sum_{j\in J}
        (\widehat z_t(j)-L_t(j))\right)\,\middle|\,\cF_{t-1}\right]\le1 .
\end{equation}
Let $Z_0=1$, and let $Z_t$ be the product of the exponential factors in~\eqref{loc:one-round-moment}
from round $1$ through round $t$. The factor $Z_{t-1}$
is nonnegative and $\cF_{t-1}$-measurable, so $\E[Z_t\mid\cF_{t-1}]\le Z_{t-1}$. Iterating gives
$\E Z_T\le1$. On the event in~\eqref{loc:ix-subset}, $Z_T>e^\Lambda$. Markov's inequality now gives
the asserted probability.

\emph{Step 3. Obtain a simultaneous bound for all fixed actions.}
Apply~\eqref{loc:ix-subset} once with $J=\{b\}$ for each $b$, and once with $J=[A]$. A union bound yields,
with failure probability at most $(A+1)e^{-\Lambda}$,
\begin{equation}\label{loc:estimate-sums}
\begin{aligned}
 \sum_{t=1}^T(\widehat z_t(b)-L_t(b))&\le\frac{\Lambda}{2\gamma}
                  &&\text{for all }b,\\
 \sum_{t=1}^T\sum_{j=1}^A\widehat z_t(j)
 &\le\sum_{t=1}^T\sum_{j=1}^AL_t(j)+\frac{\Lambda}{2\gamma}
 \le AT+\frac{\Lambda}{2\gamma}.
\end{aligned}
\end{equation}
The last inequality uses $0\le L_t(j)\le1$. Using $J=[A]$ once incurs just one confidence term for the sum.
On this event split
$\sum_t(Y_t-L_t(b))=\sum_t(Y_t-\widehat z_t(b))+\sum_t(\widehat z_t(b)-L_t(b))$.
Applying~\eqref{loc:pathwise} to the first difference and then using~\eqref{loc:estimate-sums} gives
\begin{equation}\label{loc:ix-sharper}
\begin{aligned}
 \sum_{t=1}^T(Y_t-L_t(b))
 &\le \frac{\log A}{\gamma}
      +\frac{3\gamma}{2}\left(AT+\frac{\Lambda}{2\gamma}\right)
      +\frac{\Lambda}{2\gamma}\\
 &=\frac{\log A+\Lambda/2}{\gamma}+\frac{3\gamma AT}{2}
      +\frac{3\Lambda}{4}
 \le 3\sqrt{AT\Lambda}+\Lambda .
\end{aligned}
\end{equation}
Indeed, $\log A\le\Lambda$, and our choice $\gamma=\sqrt{\Lambda/(AT)}$ makes both
$\Lambda/\gamma$ and $\gamma AT$ equal to $\sqrt{AT\Lambda}$. The bound in~\eqref{loc:ix-sharper} implies~\eqref{loc:ix-deviation} because $3\le5$.

\emph{Step 4. Control the loss under the learner's own strategy.}
For this comparison the relevant conditional mean is $\bar L_t=\E[Y_t\mid\cF_{t-1}]$. Thus
$D_t=Y_t-\bar L_t$ is $\cF_t$-measurable, has conditional mean zero given $\cF_{t-1}$, and satisfies
$|D_t|\le1$ because both terms lie in $[0,1]$. The partial sums of the $D_t$ are therefore a martingale
with increments bounded in absolute value by one. The two-sided Azuma--Hoeffding inequality~\cite[Appendix]{cesa2006prediction} for these increments gives, for $z>0$,
\[
 \Prob\!\left\{\left|\sum_{t=1}^T D_t\right|\ge z\right\}
       \le 2\exp\!\left(-\frac{z^2}{2T}\right).
\]
Choosing $z=\sqrt{2T\Lambda}$ proves~\eqref{loc:ix-obedience} except on an event of probability at most
$2e^{-\Lambda}$. The intersection with the event of Step~3 therefore has failure probability at most
$(A+3)e^{-\Lambda}\le(2A+3)e^{-\Lambda}$. On this same event both~\eqref{loc:ix-sharper}
and~\eqref{loc:ix-obedience} hold.
\end{proof}

\subsection{Proof of Theorem~\ref{loc:certificate}}\label{sec:proof-certificate}

\begin{proof}[Proof of Theorem~\ref{loc:certificate}]
We fix the continuation vectors for the duration of the local run. The proof first computes the loss
distribution that one player sees. It then converts the two loss comparisons of Lemma~\ref{loc:ix} into
payoff comparisons for the joint recommendation law returned by the algorithm.

\emph{Step 1. Identify the payoff being estimated for a fixed action.}
We realize the table of potential batches described in Section~\ref{sec:overview} on an enlarged
probability space. Given the history through round $t-1$, draw a table
$B_t=(B_t(a):a\in\cA)$ independently of the current action draws. The entries are independent,
and $B_t(a)$ consists of $n$ independent samples with distribution $p_a$.
Only $B_t(a_t)$ is revealed to the algorithm, and all players use this same entry. Given the
history and $a_t=a$, its conditional law is $p_a^{\otimes n}$, exactly that of the queried
batch. Starting from the same weights and applying this identity at each round shows by induction
that the observable transcript and the output have their original joint law. The unobserved
entries therefore account for no simulator queries.

Let $\cF_t$ include the joint action and the entire auxiliary table in each of the first $t$
rounds. The strategies for round $t$ are functions of past observations and hence are
$\cF_{t-1}$-measurable. Fix player $i$ and let $\widehat p_t^{\,a}$ denote the empirical distribution
of $B_t(a)$. If the opponents draw $a_{-i,t}$, the loss that the player would observe from action
$b$ is
\[
 z_{i,t}(b)=1-
 \frac{r_i(b,a_{-i,t})+
     \rob_{\rho_i}(\widehat p_t^{\,(b,a_{-i,t})},v_i)-c_i}{R}.
\]
Conditional on $\cF_{t-1}$, the player's action $a_{i,t}\sim x_{i,t}$ is independent of both the
opponents' product draw $a_{-i,t}\sim x_{-i,t}=\bigotimes_{j\ne i}x_{j,t}$ and the table.
The pair $(a_{-i,t},B_t)$ is jointly independent of $a_{i,t}$ given this history, so the entire
loss vector $z_{i,t}$ is conditionally independent of $a_{i,t}$. At the played action its value
is the observed loss $Y_{i,t}=z_{i,t}(a_{i,t})=\ell_{i,t}$.

For each candidate action $b$, conditioning on $a_{i,t}=b$ is legitimate because its weight is
positive. Given this event and $\cF_{t-1}$, the opponents have law $x_{-i,t}$. For each realization
$a_{-i}$, the batch used in the payoff has law $p_{(b,a_{-i})}^{\otimes n}$.
Averaging the observed payoff first over the successors gives $\bar g_i(b,a_{-i})$
by~\eqref{loc:mean-game}. Averaging next over the opponents gives the mean payoff for the fixed action $b$.
A further average over $b\sim x_{i,t}$ gives the mean payoff for playing the current strategy. In symbols,
these two payoffs are
\[
 \bar D_{i,t}(b)=\sum_{a_{-i}}x_{-i,t}(a_{-i})\,\bar g_i(b,a_{-i}),
 \qquad
 \bar C_{i,t}=\sum_a x_t(a)\,\bar g_i(a),
 \qquad x_t=\bigotimes_jx_{j,t}.
\]

\emph{Step 2. Apply the bandit bound to the observed losses.}
The algorithm feeds $Y_{i,t}=1-(\widehat g_{i,t}-c_i)/R$ to player $i$'s update. It follows from
Step~1 and this affine rescaling that $L_{i,t}(b)=\E[z_{i,t}(b)\mid\cF_{t-1}]$ equals the
conditional mean of the observed loss given $a_{i,t}=b$. More explicitly,
\begin{align}
 L_{i,t}(b)
 &=\E[Y_{i,t}\mid\cF_{t-1},a_{i,t}=b]
 =1-\frac{\bar D_{i,t}(b)-c_i}{R},\label{loc:comparator-mean}\\
 \bar L_{i,t}
 &=\E[Y_{i,t}\mid\cF_{t-1}]
 =1-\frac{\bar C_{i,t}-c_i}{R}.\label{loc:play-mean}
\end{align}
We now verify the remaining hypotheses of Lemma~\ref{loc:ix}. The pair $(z_{i,t},a_{i,t})$ is $\cF_t$-measurable. For every candidate action $b$,
\eqref{loc:centered-range} applied to its empirical row gives $0\le z_{i,t}(b)\le1$.
Together with the conditional independence proved in Step~1, these facts verify~\eqref{loc:ix-assumption}. The draw and the weight update are exactly~\eqref{loc:ix-update} with $A=A_i$ and
$\gamma_i=\sqrt{\Lambda/(A_iT)}$. The common parameter $\Lambda=\log(8mA_*/\delta)$ is positive and
at least $\log A_i$. Hence the simultaneous bounds~\eqref{loc:ix-sharper} and~\eqref{loc:ix-obedience} from the proof of the lemma apply to these conditional means. Taking a union
bound over $i$ costs at most
\[
 e^{-\Lambda}\sum_{i=1}^m(A_i+3)
 \le4mA_*e^{-\Lambda}\le\delta ,
\]
where $A_*\ge1$. The estimates for different players can use the same successors because this last
calculation only sums their failure probabilities.

\emph{Step 3. Bound each deviation against the returned recommendation law.}
Fix a realization in the simultaneous event of Step~2. The algorithm returns
$\mu=T^{-1}\sum_tx_t$. Marginalization is linear, so for each opponents' action vector $a_{-i}$ we have
$\mu_{-i}(a_{-i})=T^{-1}\sum_tx_{-i,t}(a_{-i})$. In particular, a player who commits to $b$ against the
returned law receives mean-game payoff $T^{-1}\sum_t\bar D_{i,t}(b)$. On the other hand,
$U_i=T^{-1}\sum_t\widehat g_{i,t}$ is the average payoff the player observed. Subtracting
\eqref{loc:comparator-mean} from $Y_{i,t}$ cancels the centering constants and reverses the order
of the two payoffs because the update uses losses. Thus
\[
 \E_{a_{-i}\sim\mu_{-i}}\bar g_i(b,a_{-i})-U_i
 =\frac1T\sum_{t=1}^T\bigl(\bar D_{i,t}(b)-\widehat g_{i,t}\bigr)
 =\frac{R}{T}\sum_{t=1}^T\bigl(Y_{i,t}-L_{i,t}(b)\bigr).
\]
The right-hand side is at most $R(3\sqrt{A_i\Lambda/T}+\Lambda/T)$ by~\eqref{loc:ix-sharper}.
This inequality holds for every $b$ on one event. The signed bias estimate proved in
Proposition~\ref{est:bias}, applied to $p_{(b,a_{-i})}$ and $v_i$ at every row, gives
\[
 0\le\bar g_i(b,a_{-i})-g_i(b,a_{-i})\le\frac{\spn(v_i)}{4\sqrt n}.
\]
In particular $\beta_i\le\spn(v_i)/(4\sqrt n)$ and $\beta_i=0$ at either endpoint radius
by the same proof. Replacing the mean-game payoff by the
true payoff preserves its upper bound against $\mu_{-i}$. Maximizing over $b$ gives
\[
 M_i-U_i\le R\bigl(3\sqrt{A_i\Lambda/T}+\Lambda/T\bigr).
\]
This implies~\eqref{loc:deviation-bound} since $\beta_i\ge0$ and $3\le5$.

\emph{Step 4. Evaluate obedience under the same law.}
Here the relevant distribution is the joint $\mu$. Using~\eqref{loc:play-mean} in the same loss-to-payoff
conversion gives
\[
 \E_{a\sim\mu}\bar g_i(a)-U_i
 =\frac1T\sum_{t=1}^T\bigl(\bar C_{i,t}-\widehat g_{i,t}\bigr)
 =\frac{R}{T}\sum_{t=1}^T(Y_{i,t}-\bar L_{i,t}).
\]
The absolute value of this expression is at most $R\sqrt{2\Lambda/T}$ by~\eqref{loc:ix-obedience}.
Using $C_i=\E_{a\sim\mu}g_i(a)$ now gives
\[
 C_i-U_i=\bigl(\E_{a\sim\mu}\bar g_i(a)-U_i\bigr)
                  -\sum_a\mu(a)\bigl(\bar g_i(a)-g_i(a)\bigr).
\]
Every term in the weighted average of biases belongs to $[0,\beta_i]$. The same bound therefore holds
for the average, even though $\mu$ was learned from these samples. Indeed, this is a pointwise inequality
for each row and the weights sum to one on every realization. The triangle inequality now
gives~\eqref{loc:obedience-bound}.

Finally, the player-specific range in Lemma~\ref{est:range}(a), applied to the selected empirical
row, gives $c_i\le\widehat g_{i,t}\le c_i+1+(1-\rho_i)\spn(v_i)$ at each round. The same endpoints
therefore bound the average $U_i$ on every run. There is one simulator batch per round,
shared by the $m$ updates, so the number of transition queries is exactly $nT$.
\end{proof}

\begin{remark}[Inputs chosen from past data]\label{loc:conditional}
The same probability bound holds when a local call takes inputs computed from earlier observations.
Let $\mathcal G$ be the information just before the call, and suppose its inputs are
$\mathcal G$-measurable and satisfy the hypotheses of Theorem~\ref{loc:certificate} almost surely.
Conditional on $\mathcal G$, these inputs are fixed. Include $\mathcal G$ in the initial filtration
$\cF_0$ of the proof above. Given each subsequent pre-round history the action draw still has the
specified product law. Given the selected row the simulator supplies $n$ fresh draws, so the
potential batches of Step~1 in the proof of Theorem~\ref{loc:certificate} can again be realized
independently of the current action draws.
Consequently~\eqref{loc:comparator-mean} and~\eqref{loc:play-mean} hold in this filtration.
The probability that at least one player's certificate fails, conditional on $\mathcal G$, is therefore
at most the chosen local failure probability.
\end{remark}

\section{Proof of Theorem~\ref{thm:simple-global}}\label{app:global}

This appendix combines the local guarantee of Appendix~\ref{sec:local} with the backward recursion. The argument needs two further facts. Lemma~\ref{glob:range} shows that the continuation vectors produced by Algorithm~\ref{alg:backward} have the spans assumed by the parameter choice, so that the local runs satisfy the hypotheses of Theorem~\ref{loc:certificate}. Lemma~\ref{glob:propagation} shows how the errors of the local certificates accumulate over the stages. The proof of the theorem then assembles these pieces.

\subsection{The spans of the learned continuation vectors}

\begin{lemma}[Span of the learned continuation vectors]\label{glob:range}
On every run of Algorithm~\ref{alg:backward},
\[
 \spn(U_{i,h})\le d_{i,h}
 \qquad\text{for all }i\in[m]\text{ and }h\in[H+1],
\]
where $d_{i,h}$ is defined in~\eqref{glob:range-recursion}. Consequently, for every stage $h<H$, the local runs at stage $h$ satisfy the range condition~\eqref{loc:range-condition} with $R=d_h$.
\end{lemma}

\begin{proof}
We prove the bound by backward induction for a fixed player $i$. The bound is deterministic because the
only fact we use about a batch is that its empirical row is a probability distribution.
At stage $H+1$ the vector is zero and $d_{i,H+1}=0$. At stage $H$, $U_{i,H}(s)$ is an expectation of
rewards in $[0,1]$ for every state $s$. All its entries therefore lie in the same interval $[0,1]$ and its
span is at most $1=d_{i,H}$.

For the induction step, suppose $\spn(U_{i,h+1})\le d_{i,h+1}$ and $h<H$. Write $v_i=U_{i,h+1}$ and
$c_i=\min_{s'}v_i(s')$. The local games at all states of stage $h$ use this \emph{same} continuation vector.
For any state $s$, any joint action $a$, and any empirical row $\widehat p$ returned there,
Lemma~\ref{est:range} gives
\[
 c_i\le r_{i,h}(s,a)+\rob_{\rho_i}(\widehat p,v_i)
      \le c_i+1+(1-\rho_i)\spn(v_i)
      \le c_i+d_{i,h}.
\]
The last step uses the induction hypothesis, $1-\rho_i\ge0$, and the recursion
$d_{i,h}=1+(1-\rho_i)d_{i,h+1}$. Every $U_{i,h}(s)$ averages quantities in this interval. Its
endpoints are independent of the current state $s$, so
$\max_s U_{i,h}(s)-\min_s U_{i,h}(s)\le(c_i+d_{i,h})-c_i=d_{i,h}$.
This completes the induction for each player. In particular,
$1+(1-\rho_i)\spn(U_{i,h+1})\le d_{i,h}\le d_h$ for every $i$, exactly the common bound on $R$ needed
in~\eqref{loc:range-condition}.
\end{proof}

\subsection{Propagation of the local errors}

Recall the quantities $C_{i,h}(s)$ and $M_{i,h}(s)$ defined in~\eqref{glob:obedience-backup} and~\eqref{glob:deviation-backup}. They are the right-hand sides of the robust Bellman equations~\eqref{eq:obey} and~\eqref{eq:br} with the vector $U_{i,h+1}$ in place of the true continuation values $V^{\pi}_{i,h+1}$ and $V^{\dagger,\pi_{-i}}_{i,h+1}$. The local certificates compare $U_{i,h}$ with $C_{i,h}$ and $M_{i,h}$, whereas the equilibrium gap involves the true values. The next lemma bridges this difference. It is a deterministic statement about an arbitrary policy and arbitrary vectors.

\begin{lemma}[Propagation of the local errors]\label{glob:propagation}
Let $\pi$ be a Markov joint policy, and for every player~$i$ let $U_{i,1},\dots,U_{i,H+1}\in\mathbb R^{\cS}$ be vectors with $U_{i,H+1}=0$. Define $C_{i,h}$ and $M_{i,h}$ by~\eqref{glob:obedience-backup} and~\eqref{glob:deviation-backup}. Suppose that there are numbers $\kappa_1,\dots,\kappa_H\ge0$ such that, for all $i\in[m]$, $h\in[H]$, and $s\in\cS$,
\begin{equation}\label{glob:abstract-certificate}
 U_{i,h}(s)-C_{i,h}(s)\le\kappa_h
 \qquad\text{and}\qquad
 M_{i,h}(s)-U_{i,h}(s)\le\kappa_h .
\end{equation}
Then, for all $i\in[m]$, $h\in[H+1]$, and $s\in\cS$,
\begin{align}
 U_{i,h}(s)-V_{i,h}^{\pi}(s)&\le\sum_{j=h}^H\kappa_j,
       \label{glob:obey-propagated}\\
 V_{i,h}^{\dagger,\pi_{-i}}(s)-U_{i,h}(s)
       &\le\sum_{j=h}^H\kappa_j,\label{glob:br-propagated}
\end{align}
and consequently
\[
 \max_{s\in\cS}\CEgap(\pi,s)\le2\sum_{h=1}^H\kappa_h .
\]
\end{lemma}

\begin{proof}
The certificates in~\eqref{glob:abstract-certificate} compare payoffs computed from the learned vector
$U_{i,h+1}$. The two values in the equilibrium criterion instead use the continuation value of obedience
and the continuation value of a best response, respectively. We first show how replacing one continuation
vector by another changes each backup. We can then compare the vectors by backward induction, with one
error for each remaining stage.

Write $K_h=\sum_{j=h}^H\kappa_j$ with $K_{H+1}=0$, so $K_h=\kappa_h+K_{h+1}$.
At a fixed $(h,s)$ let $p_a=P_h(\cdot\mid s,a)$. Fix $i$ and $\pi$, and denote by $B_h^{\mathrm{o}}$ and $B_h^{\mathrm{d}}$ the two maps on continuation
vectors given by the right-hand sides of~\eqref{eq:obey} and~\eqref{eq:br}. The policy used to form the
action averages is fixed in both maps. By definition,
$B_h^{\mathrm{o}}U_{i,h+1}=C_{i,h}$ and $B_h^{\mathrm{d}}U_{i,h+1}=M_{i,h}$.
If $u\le v+c\one$, then~\eqref{glob:one-sided-lipschitz} gives
$\rob_{\rho_i}(p_a,u)\le\rob_{\rho_i}(p_a,v)+c$ separately for every joint action $a$.
Adding the same reward and averaging against a probability law preserves this bound.
For a deviation we first apply it at each fixed own action $b$. Since the right-hand error $c$ is the
same for every $b$, taking the maximum preserves it as well. We have established that
\[
 B_h^{\mathrm{o}}u\le B_h^{\mathrm{o}}v+c\one,\qquad
 B_h^{\mathrm{d}}u\le B_h^{\mathrm{d}}v+c\one .
\]
In particular a one-sided error in a continuation vector passes through either backup with coefficient one.

For compactness in the induction, write $W_h=V^\pi_{i,h}$ and $W_h^\dagger=V_{i,h}^{\dagger,\pi_{-i}}$ for the
two \emph{value vectors}. Lemma~\ref{lem:bellman} applies because $\pi$ is a Markov recommendation policy
in the rectangular game of Section~\ref{sec:model}. It gives $W_h=B_h^{\mathrm{o}}W_{h+1}$ and
$W_h^\dagger=B_h^{\mathrm{d}}W_{h+1}^\dagger$, including the best response over all admissible coarse deviations.
At $H+1$, $U_{i,H+1}=W_{H+1}=W_{H+1}^\dagger=0$ and $K_{H+1}=0$. Assume at $h+1$ that
$U_{i,h+1}\le W_{h+1}+K_{h+1}\one$ and $W_{h+1}^\dagger\le U_{i,h+1}+K_{h+1}\one$.

For obedience the current certificate bounds $U_{i,h}$ from above. Applying it first and then inserting the induction hypothesis into the obedience backup gives
\[
 U_{i,h}
 \le C_{i,h}+\kappa_h\one
 = B_h^{\mathrm{o}}U_{i,h+1}+\kappa_h\one
 \le B_h^{\mathrm{o}}W_{h+1}+(K_{h+1}+\kappa_h)\one
 = W_h+K_h\one .
\]
For deviations the induction hypothesis bounds the true best-response continuation from above by
the learned continuation plus $K_{h+1}\one$. We insert it into the deviation backup and then apply the current certificate to obtain
\[
 W_h^\dagger=B_h^{\mathrm{d}}W_{h+1}^\dagger
 \le B_h^{\mathrm{d}}U_{i,h+1}+K_{h+1}\one
 =M_{i,h}+K_{h+1}\one
 \le U_{i,h}+K_h\one .
\]
In both lines $K_h=\kappa_h+K_{h+1}$. This proves the two one-sided bounds by induction. At any
state $s$, the difference $W_h^\dagger(s)-W_h(s)$ is the sum of $W_h^\dagger(s)-U_{i,h}(s)$ and
$U_{i,h}(s)-W_h(s)$. Each is at most $K_h$, so the positive deviation gain is at most $2K_h$. Taking $h=1$ and maximizing over players
and initial states gives $\max_s\CEgap(\pi,s)\le2K_1$, as claimed.

Finally, the stronger local condition $|U_{i,h}-C_{i,h}|\le\kappa_h\one$ also gives
$C_{i,h}\le U_{i,h}+\kappa_h\one$, coordinatewise. To bound $W_h$ in the reverse direction, start
again at $H+1$ and assume $W_{h+1}\le U_{i,h+1}+K_{h+1}\one$. The obedience backup then gives
\[
 W_h=B_h^{\mathrm{o}}W_{h+1}
 \le B_h^{\mathrm{o}}U_{i,h+1}+K_{h+1}\one
 = C_{i,h}+K_{h+1}\one
 \le U_{i,h}+K_h\one .
\]
Together with~\eqref{glob:obey-propagated}, this proves the additional implication
$|U_{i,h}(s)-V^\pi_{i,h}(s)|\le K_h$ when the local obedience estimates are two-sided.
\end{proof}

\subsection{Proof of the theorem}

\begin{proof}[Proof of Theorem~\ref{thm:simple-global}]
\emph{Step 1. Check the local guarantee at a data-dependent continuation.}
Assume $H\ge2$, as in the theorem. Consider the call at a particular pair $(h,s)$ with $h<H$, and
let $\mathcal G_{h,s}$ be all information accumulated before that call. It includes the completed
training at stage $h+1$ and, if applicable, the earlier calls at other states of stage $h$.
Thus the input $v_i=U_{i,h+1}$ is fixed conditional on $\mathcal G_{h,s}$. The nominal kernel and
rewards are fixed parameters of the game. The random action draws in the new call are fresh, and
given each chosen row the simulator uses fresh randomness at each of its $n_h$ queries. These
are the conditional sampling laws used in Theorem~\ref{loc:certificate}, as detailed in
Remark~\ref{loc:conditional}.

Write $\delta'=\delta/[S(H-1)]$. Since $S\ge1$, $H\ge2$ and $0<\delta<1$, we have
$0<\delta'<1$. We apply Corollary~\ref{loc:accuracy} with local confidence $1-\delta'$,
range $R=d_h$ and accuracy $\epsz=\varepsilon/(2H)$. The local parameter requirements are satisfied since
\begin{align*}
 \max_i\{1+(1-\rho_i)\spn(v_i)\}&\le d_h=R,
     &0<\epsz&\le1\le d_h,\\
 n_h&\ge\frac{d_{h+1}^2}{\epsz^2}
       \ge\frac{\max_i\spn(v_i)^2}{\epsz^2},
     &T_h&\ge\frac{256 A_*R^2\Lambda}{\epsz^2}.
\end{align*}
The span inequalities hold on every training sample path by Lemma~\ref{glob:range}. The remaining
inequalities follow from the ceilings in~\eqref{glob:parameters}, $d_h\ge1$, and $\varepsilon\le1$.
Finally, the global choice of $\Lambda$ equals the required local logarithm
$\log(8mA_*/\delta')$. Thus on each pre-call history the corollary applies to all players simultaneously.

\emph{Step 2. Obtain one event for all calls.}
Let $E_{h,s}$ be the event that the returned law and value estimates satisfy
\[
 |U_{i,h}(s)-C_{i,h}(s)|\le\epsz,\qquad
 M_{i,h}(s)-U_{i,h}(s)\le\epsz\qquad\text{for every }i.
\]
The quantities $C_{i,h}(s)$ and $M_{i,h}(s)$ here are the exact backups for the law returned by
this call, evaluated at its input $U_{i,h+1}$. Step~1 gives
$\Prob(E_{h,s}^{\,c}\mid\mathcal G_{h,s})\le\delta'$ almost surely. The tower property applied to
the indicator of failure yields
$\Prob(E_{h,s}^{\,c})=\E[\Prob(E_{h,s}^{\,c}\mid\mathcal G_{h,s})]\le\delta'$.
A union bound over the $S(H-1)$ calls therefore gives
\[
 \Prob\!\left(\bigcap_{h<H,\,s\in\cS}E_{h,s}\right)
 \ge 1-S(H-1)\delta'=1-\delta .
\]
Earlier calls can affect later inputs. The conditional guarantee is what allows the same calculation
for each call despite this dependence.

\emph{Step 3. Pass from the certificates to the equilibrium guarantee.}
It remains to supply the last stage for Lemma~\ref{glob:propagation}. When $h=H$,
$U_{i,H+1}=0$ makes every continuation backup zero. To see that the terminal feasibility problem has a solution, observe that there are finitely many players and nonempty finite action sets with finite payoffs, so Nash's
finite-game theorem gives a mixed Nash equilibrium~\cite{nash_non-cooperative_1951}. Under
its product law a player's expected equilibrium payoff is at least its payoff from any fixed
action against the opponents' marginal. These are precisely the linear feasibility inequalities
in~\eqref{glob:terminal}. The algorithm may take any feasible law $\nu=\pi_H(\cdot\mid s)$.
For this law the terminal estimates satisfy
\[
 C_{i,H}(s)=\E_{a\sim\nu}r_{i,H}(s,a)=U_{i,H}(s),\qquad
 M_{i,H}(s)=\max_b\E_{a_{-i}\sim\nu_{-i}}r_{i,H}(s,(b,a_{-i}))\le U_{i,H}(s).
\]

Now fix any training realization on the event of Step~2. Its pair $(\pi,U)$ satisfies all the hypotheses
of Lemma~\ref{glob:propagation} with $\kappa_H=0$ and $\kappa_j=\epsz$ for $j<H$. The lemma is
deterministic, so it applies to this realized policy even though the policy and the value estimates
were learned from the same observations. Since $K_h=\sum_{j=h}^H\kappa_j=(H-h)\epsz$, we obtain
\[
 \bigl(V^{\dagger,\pi_{-i}}_{i,h}(s)-V^\pi_{i,h}(s)\bigr)_+
 \le 2(H-h)\epsz\qquad\text{for every }i,h,s.
\]
In particular the right-hand side at $h=1$ is
$2(H-1)\epsz=(H-1)\varepsilon/H\le\varepsilon$. Taking the maximum over players and initial
states proves part~(i) of the theorem on the event already shown to have probability at least $1-\delta$.

\emph{Step 4. Count the transition queries.}
The terminal solve uses known rewards. At each earlier pair $(h,s)$, all players share the same $n_h$
successors in a round and the algorithm runs exactly $T_h$ rounds. Summing over the pairs gives the
identity $N=S\sum_{h=1}^{H-1}n_hT_h$ in~\eqref{glob:exact-budget} on every execution.

For $h<H$, both $d_h$ and $d_{h+1}$ are at least one. Since $\epsz\le1$ and
$\Lambda=\log(8mA_*/\delta')\ge\log 8>1$, the numbers
$d_{h+1}^2/\epsz^2$ and $A_*d_h^2\Lambda/\epsz^2$ are each at least one. Therefore rounding up the
batch size multiplies it by at most two, and rounding up $256 A_*d_h^2\Lambda/\epsz^2$ gives at
most $257 A_*d_h^2\Lambda/\epsz^2$. Substituting these estimates in the exact count yields
\[
 N\le 2\cdot257\,\frac{S A_*\Lambda}{\epsz^4}
                  \sum_{h=1}^{H-1}d_h^2d_{h+1}^2
  =8224\,\frac{S A_*H^4\Lambda}{\varepsilon^4}
                  \sum_{h=1}^{H-1}d_h^2d_{h+1}^2 ,
\]
where $\epsz^{-4}=16H^4/\varepsilon^4$. The geometric sum defining $d_{i,h}$ in
\eqref{glob:range-recursion} is at most $H-h+1$. For $\rho_i>0$ it is also at most $1/\rho_i$,
and for $\rho_i=0$ the former bound applies. Hence $d_h,d_{h+1}\le D$ for every summand,
so the sum is at most $(H-1)D^4\le HD^4$. Together with $\Lambda\le\log(8mA_*SH/\delta)$
this proves the second inequality in~\eqref{glob:refined-budget} and completes part~(ii).
\end{proof}

\begin{remark}[Degenerate cases]\label{rem:degenerate}
When $H=1$, a CCE of the known reward game at each state is already an exact robust CCE.
For $H\ge2$, the endpoint formulas of Lemma~\ref{est:tail} give a useful modification when every
$\rho_i$ is either zero or one. At radius zero, $r_i(a)+v_i(X)$ for one draw $X\sim p_a$ has mean
$r_i(a)+p_av_i=g_i(a)$. At radius one the empirical backup of any row is the deterministic
number $\min_{s'}v_i(s')$. Thus a single shared successor gives $\bar g_i=g_i$ for every player
even when the endpoint radii differ between players.

Set $n_h=1$ and retain $T_h$ from~\eqref{glob:parameters}. To justify this change, apply
Theorem~\ref{loc:certificate} directly. It allows every positive batch size, and now has $\beta_i=0$.
The range bound is still pathwise and the two terms depending on $T_h$ satisfy the same bounds as in
the proof of Corollary~\ref{loc:accuracy}. We obtain the same local accuracy $\epsz$ and the same
probability for each call, so the simultaneous-event and propagation arguments above apply.
The number of queries is now
\[
 S\sum_{h=1}^{H-1}T_h
 \le \frac{257 S A_*\Lambda}{\epsz^2}
       \sum_{h=1}^{H-1}d_h^2
 \le \frac{257 S A_*\Lambda H^3}{\epsz^2}
 =\widetilde O(SA_*H^5/\varepsilon^2).
\]
Here $d_h\le H-h+1$ and there are at most $H$ summands, each at most $H^2$.

If all radii equal one, an exact solution uses zero transition queries. Choose a CCE of the known reward
game at every $(h,s)$. Starting from $U_{i,H+1}=0$, define
$U_{i,h}(s)=\E_{a\sim\pi_h(\cdot\mid s)}r_{i,h}(s,a)+\min_{s'}U_{i,h+1}(s')$.
The second term is common to every joint action at $(h,s)$, so it cancels from each unilateral incentive
inequality. Thus $C_{i,h}=U_{i,h}$ and $M_{i,h}\le U_{i,h}$ at every stage. Applying
Lemma~\ref{glob:propagation} with all $\kappa_h=0$ proves exactness.
\end{remark}

\begin{remark}[Avoiding the terminal linear program]\label{rem:terminal-bandit}
If the known reward table is accessed one joint action at a time, we can obtain the terminal
recommendation law by a reward-only run of Algorithm~\ref{alg:local}. Set $\epsz=\varepsilon/(2H)$ and
use $\Lambda=\log(8mA_*SH/\delta)$ at every stage. At a terminal state take $v_i=0$, $R=1$ and
$T_H=\lceil256A_*\Lambda/\epsz^2\rceil$. The update observes $\widehat g_{i,t}=r_{i,H}(s,a_t)$ and
returns the same payoff averages and mixture of strategies as the local algorithm.

To apply Theorem~\ref{loc:certificate}, observe that the terminal row backup of the zero vector is zero
for both the true row and every possible empirical row. Thus the reward-only updates have precisely
the same law as the local updates with $v_i=0$ and a batch of size one whose successor is ignored.
In Algorithm~\ref{alg:local} this skips the query in line~5 and uses the known reward in line~7. The loss lies in $[0,1]$ because the
reward does, and the local bias is zero. The chosen $T_H$ gives each of the two local errors at most
$\epsz$ by the calculation following~\eqref{loc:accuracy-parameters}. The recorded terminal averages
lie in $[0,1]$ on every run, which is the only terminal property used to start the span induction.
At earlier stages we can therefore use the same $d_h,n_h$ and the round counts in~\eqref{glob:parameters}
with the updated $\Lambda$.

There are now $SH$ local events, each with conditional failure probability at most $\delta/(SH)$.
Their union bound gives total failure probability at most $\delta$. At the terminal stage the certificate
has accuracy $\epsz$ as well, so Lemma~\ref{glob:propagation} applies with $\kappa_h=\epsz$ at all $H$
stages. The initial gap is at most $2H\epsz=\varepsilon$. The only transition queries are those made at
$h<H$, with the updated logarithmic factor.
\end{remark}

\section{Robust dynamic programming: proof of Lemma~\ref{lem:bellman}}\label{app:bellman}

Lemma~\ref{lem:bellman} states that the robust value of obeying and the best-response value satisfy the robust Bellman equations~\eqref{eq:obey} and~\eqref{eq:br}. The argument is the classical one for rectangular uncertainty sets~\cite{iyengar2005robust,nilim2005robust}. We include it because the best-response value~\eqref{eq:br-value} is defined with a supremum over history-dependent deviations and a minimum over kernels fixed in advance, and the proof must show that neither feature changes the recursion.

\begin{proof}[Proof of Lemma~\ref{lem:bellman}]
Fix a player $i$ and a Markov joint policy $\pi$. We use two kernels. One attains the obedience
value of $\pi$ from every starting pair. The other bounds every coarse deviation against $\pi_{-i}$
at once, which lets us compare history-dependent deviations with a single Markov best response.

\emph{Step 1. Construct the candidate values and kernels.}
Write $p_{k,s,a}=P_k(\cdot\mid s,a)$. Starting from $W_{H+1}=B_{H+1}=0$, define $W_k$ by the
obedience recursion~\eqref{eq:obey} with continuation $W_{k+1}$, and $B_k$ by the best-response
recursion~\eqref{eq:br} with continuation $B_{k+1}$. At this point $W_k$ and $B_k$ are just the
vectors computed by the recursions. We will identify them with the two worst-case returns in
\eqref{eq:robust-value} and~\eqref{eq:br-value}.

For either sequence $F\in\{W,B\}$, consider the minimization of $qF_{k+1}$ over
$\cP_{i,k}(s,a)$. The ball contains $p_{k,s,a}$ and is a closed subset of the finite-dimensional
probability simplex, so it is nonempty and compact, including at $\rho_i=0$. The objective is linear
and continuous. The extreme-value theorem therefore supplies a minimizing row at every index.
Since $\cP_i$ is the product over $(k,s,a)$, all the chosen
rows belong to one admissible kernel $Q^F$, satisfying
\[
 Q^F_k(\cdot\mid s,a)F_{k+1}
       =\rob_{\rho_i}(p_{k,s,a},F_{k+1})
       \qquad\text{for every }(k,s,a).
\]
The row for a given action only needs to minimize the designated continuation. In particular
$Q^W$ and $Q^B$ may differ. Each is a fixed, stage-dependent kernel chosen before play.

\emph{Step 2. Reduce a return comparison to one-step comparisons.}
Start the game at an arbitrary pair $(h,s)$. Let $S_k,A_k$ be the state and realized joint action,
and let $\mathcal H_k$ be the information from play up to $S_k$, before the stage-$k$
recommendations and actions. For either candidate $F$, define the one-step difference
\[
 \Delta_k(F)=r_{i,k}(S_k,A_k)+F_{k+1}(S_{k+1})-F_k(S_k).
\]
On every path $\sum_{k=h}^H(F_{k+1}(S_{k+1})-F_k(S_k))=-F_h(s)$, because the intermediate
terms cancel and the terminal vector is zero. For any fixed behavior $\tau$ and kernel $Q$, write $\E_{h,s}^{\tau,Q}$ for expectation
under their trajectory law from $(h,s)$. Taking expectations therefore gives
\begin{equation}\label{bell:nominal}
 V^{\tau,Q}_{i,h}(s)-F_h(s)=
     \E_{h,s}^{\tau,Q}\sum_{k=h}^H\Delta_k(F)
   =\sum_{k=h}^H\E_{h,s}^{\tau,Q}
        \bigl[\E_{h,s}^{\tau,Q}[\Delta_k(F)\mid\mathcal H_k]\bigr].
\end{equation}
The rewards and candidate vectors are bounded on this finite horizon, so all conditional expectations
exist and the tower property applies. Thus the sign of each conditional one-step difference determines
the direction of the desired return bound.

\emph{Step 3. Verify obedience and simultaneous attainment.}
Suppose all players obey. On a history with $S_k=x$, the mediator draws $A_k$ with law
$\pi_k(\cdot\mid x)$. Given $A_k=a$, the next-state law under $Q$ is $Q_k(\cdot\mid x,a)$.
Consequently, for every $Q\in\cP_i$,
\begin{align*}
 \E_{h,s}^{\pi,Q}[\Delta_k(W)\mid\mathcal H_k]
 &=\sum_a\pi_k(a\mid x)
       \bigl[r_{i,k}(x,a)+Q_k(\cdot\mid x,a)W_{k+1}\bigr]-W_k(x)\\
 &\ge\sum_a\pi_k(a\mid x)
       \bigl[r_{i,k}(x,a)+\rob_{\rho_i}(p_{k,x,a},W_{k+1})\bigr]-W_k(x)=0 .
\end{align*}
Each feasible row gives at least its minimum, and the last equality is the definition of $W_k$.
For $Q=Q^W$, every row realizes its minimum, making the display an equality on every history.
Using~\eqref{bell:nominal} first for arbitrary $Q$ and then for $Q^W$ yields
\[
 V_{i,h}^{\pi,Q}(s)\ge W_h(s)=V_{i,h}^{\pi,Q^W}(s).
\]
This identifies $W_h(s)$ with the minimum in~\eqref{eq:robust-value}. The same $Q^W$ attains it
from every $(h,s)$ because its rows were selected at every stage and state in Step~1.

\emph{Step 4. Bound all coarse deviations under one kernel.}
For $F=B$, denote the payoff of fixing the own action $b$ in the candidate recursion by
\[
 d_k(x,b)=\sum_{a_{-i}}\pi_{-i,k}(a_{-i}\mid x)
     \bigl[r_{i,k}(x,(b,a_{-i}))
         +\rob_{\rho_i}(p_{k,x,(b,a_{-i})},B_{k+1})\bigr].
\]
Thus $B_k(x)=\max_b d_k(x,b)$. Fix any history-dependent, possibly randomized coarse deviation
$\sigma_i$, and use the fixed kernel $Q^B$. Conditional on $\mathcal H_k$ with $S_k=x$ and on the
deviator choosing $A_{i,k}=b$ of positive conditional probability, the opponents' current actions
have law $\pi_{-i,k}(\cdot\mid x)$. Indeed, their recommendations use fresh randomness at this stage,
and the deviator chooses $b$ without its current recommendation. This is exactly where the information
available to a coarse deviation enters the proof.

By construction the row of $Q^B$ gives the robust continuation for \emph{each realized joint action}.
Averaging over the opponents therefore gives
\[
 \E_{h,s}^{\sigma_i\times\pi_{-i},Q^B}
       [\Delta_k(B)\mid\mathcal H_k,A_{i,k}=b]=d_k(x,b)-B_k(x)\le0 .
\]
The inequality holds at every possible own choice. Averaging it under the deviator's conditional
action distribution and summing by~\eqref{bell:nominal} gives, for every $\sigma_i$,
\[
 \min_{Q\in\cP_i}V_{i,h}^{\sigma_i\times\pi_{-i},Q}(s)
 \le V_{i,h}^{\sigma_i\times\pi_{-i},Q^B}(s)\le B_h(s).
\]
This bound is uniform over deviations. Past observations can affect which $b$ the deviator picks,
but all these actions face the same upper bound at the current state.

\emph{Step 5. Exhibit a deviation attaining the bound.}
At each $(k,x)$ choose $b^\star_k(x)\in\arg\max_b d_k(x,b)$, which exists because $\cA_i$ is
finite and nonempty. These choices form a single deterministic Markov policy $\sigma_i^\star$.
It is an admissible coarse deviation. Against any fixed $Q\in\cP_i$, feasibility of the realized
row gives
\[
 Q_k(\cdot\mid x,(b^\star_k(x),a_{-i}))B_{k+1}
 \ge\rob_{\rho_i}(p_{k,x,(b^\star_k(x),a_{-i})},B_{k+1}).
\]
Averaging against the opponents' marginal now gives conditional expected reward plus continuation
at least $d_k(x,b^\star_k(x))=B_k(x)$. Thus the conditional difference in~\eqref{bell:nominal}
is nonnegative at each stage, and
$V_{i,h}^{\sigma_i^\star\times\pi_{-i},Q}(s)\ge B_h(s)$ for every $Q\in\cP_i$.
Taking first the minimum in this lower bound, and the supremum over deviations in the upper bound
of Step~4, gives
\[
 B_h(s)\le\min_{Q\in\cP_i}V_{i,h}^{\sigma_i^\star\times\pi_{-i},Q}(s)
 \le\sup_{\sigma_i}\min_{Q\in\cP_i}V_{i,h}^{\sigma_i\times\pi_{-i},Q}(s)
 \le B_h(s).
\]
All three quantities coincide. This proves~\eqref{eq:br}, and the same $\sigma_i^\star$ attains the
supremum at every starting pair. The argument establishes the stated order of optimization directly.

Finally, suppose nature can choose a feasible row after observing the history and current joint action.
The lower bounds in Steps~3 and~5 still hold conditionally on that information, because every selected
row belongs to the same ball at the realized index. Conversely, nature can always use the fixed kernel
$Q^W$ to attain the obedience minimum, or $Q^B$ to enforce the upper bound on every deviation.
Hence obedience and the optimal coarse-deviation value have the same recursions under this extension.
\end{proof}

\end{document}